\documentclass[10pt,twocolumn,letterpaper]{article}

\usepackage{iccv}
\usepackage{times}
\usepackage{epsfig}
\usepackage{graphicx}
\usepackage{amsmath}
\usepackage{amssymb}

\usepackage{booktabs}
\usepackage{tablefootnote}
\usepackage{multirow}
\usepackage{comment}
\usepackage{tablefootnote}
\usepackage{nicefrac}       
\usepackage{microtype}      

\usepackage[pagebackref=true,breaklinks=true,letterpaper=true,colorlinks,bookmarks=false]{hyperref}

\iccvfinalcopy 

\def\iccvPaperID{4284} 

\begin{document}

\title{iPOKE: Poking a Still Image for Controlled Stochastic Video Synthesis}

\author{
Andreas Blattmann$^{1,2}$  \quad Timo Milbich$^{1,2}$ \quad Michael Dorkenwald$^{2}$  \quad Bj{\"o}rn Ommer$^{1,2}$\\
$^1$Ludwig Maximilian University of Munich \quad $^2$IWR, Heidelberg University, Germany
}

\maketitle

\begin{abstract}
How would a static scene react to a local poke? What are the effects on other parts of an object if you could locally push it? There will be distinctive movement, despite evident variations caused by the stochastic nature of our world. These outcomes are governed by the characteristic kinematics of objects that dictate their overall motion caused by a local interaction. Conversely, the movement of an object provides crucial information about its underlying distinctive kinematics and the interdependencies between its parts. This two-way relation motivates learning a bijective mapping between object kinematics and plausible future image sequences. Therefore, we propose iPOKE -- invertible Prediction of Object Kinematics -- that, conditioned on an initial frame and a local poke, allows to sample object kinematics and establishes a one-to-one correspondence to the corresponding plausible videos, thereby providing a controlled stochastic video synthesis.
In contrast to previous works, we do not generate arbitrary realistic videos, but provide efficient control of movements, while still capturing the stochastic nature of our environment and the diversity of plausible outcomes it entails. Moreover, our approach can transfer kinematics onto novel object instances and is not confined to particular object classes. Our project page is available at \url{https://bit.ly/3dJN4Lf}.
\end{abstract}
\begin{figure}[t]
\begin{center}
\includegraphics[width=\linewidth]{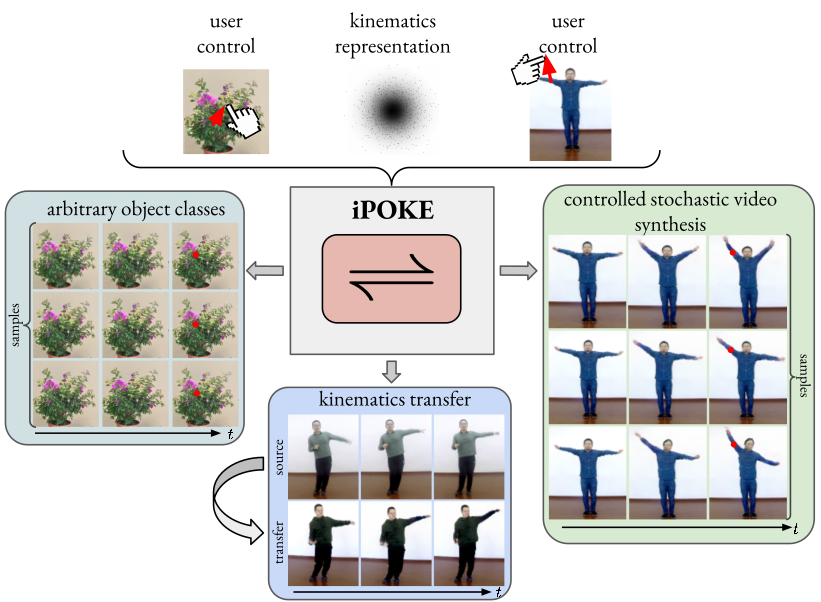}
\end{center}
   \caption{\textit{iPOKE:} Conditioned on a local poke controlling desired object motion in a static image, our invertible model learns a representation of the remaining object kinematics for arbitrary object classes. Once learned, our framework allows users to locally control intended movements while sampling diverse realistic motion for the remainder of the object and to even transfer kinematics to unseen object instances.}
\label{fig:first_page}
\vspace{-4mm}
\end{figure}
\section{Introduction}
\enlargethispage{\baselineskip}
\vspace{-2mm}
Imagine a 3-year-old standing next to a stacked pyramid of glasses in a shop. Can you sense the urge to pull one glass out---just to observe what happens. We have an inborn curiosity to understand how the world around us reacts to our actions, so we can eventually imagine and predict their outcome beforehand. This ability to predict is the prerequisite for targeted, goal-oriented interaction with our world rather than random manipulation of our environment. Once we are older, we have also learned to generalize and predict the dynamics of previously unseen objects when they are pulled or poked; and the less audacious have understood that it is often more effective to have others do daring experiments like the one above (and pay the bill) while they are learning by merely watching the outcome. While such experiments are not just fun to watch, they also help to imagine the many possible outcomes caused by the stochastic nature of the many factors beyond our control.
\\
Given a single static image, how can an artificial vision system imagine, i.e. synthesize, the many possible outcomes when locally manipulating the scene? It needs to learn how a local poke affects different parts of an object and the resulting kinematics \cite{macgregor1902elementary}. Conditioned on only the start frame and the displacement of a single pixel, we want to synthesize multiple videos, each showing the different plausible future dynamics. To render this generative, stochastic approach widely applicable, training should only require videos of objects in motion, but no ground truth information regarding the forces acting on an object such as a local poke. The representation of the kinematics should then generalize to similar objects not seen during training in contrast to instance specific models \cite{davis2015}. Moreover, the method should work for arbitrary objects, rather than being tuned to just a single class \cite{aberman2020unpaired,hbugen}. Therefore, no prior motion model is available, but all kinematics have to be learned from the unannotated video data.
\noindent
Previous work on video synthesis has mainly explored two opposing research directions: \textit{(i)} uncontrolled future frame prediction \cite{Franceschi2020, vrnn-hier, minderer2020unsupervised, sdcnet18} synthesizing videos based on a start frame, but with no control of scene dynamics, and \textit{(ii)} densely controlled video synthesis \cite{mallya2020world,vid2vid,wu2020future,Wang_2019_ICCV} demanding tedious, per-pixel guidance how the video will evolve such as by requiring the object motion to be provided per pixel \cite{mallya2020world,vid2vid,wu2020future} or a future target frame \cite{Wang_2019_ICCV}. Our sparsely controlled video synthesis based on few local user interactions constitutes the rarely investigated midground in between, allowing for specific but still efficient control of kinematics.
\\
In this paper, we present a model for exercising local control over the kinematics of objects observed in an image. Indicating movements of individual object parts with a simple mouse drag provides sufficient input for our model to synthesize plausible, holistic object motion. To capture the ambiguity in the global object articulation, we learn a dedicated latent kinematic representation. The synthesis problem is then formulated as an invertible mapping between object kinematics and video sequences conditioned on the observed object manipulation. Due to its stochastic nature, our latent representation allows to sample and transfer diverse kinematic realizations fitting to the sparse local user input to then infer and synthesize plausible video sequences as shown in Fig.~\ref{fig:first_page}.
\\
To evaluate our model on controlled stochastic video synthesis, we conduct quantitative and qualitative experiments on four different datasets exhibiting complex and highly articulated objects, such as humans and plants. Comparisons with the state-of-the-art in stochastic and controlled video prediction demonstrate the capability of our model to predict and synthesize plausible, diverse object articulations inferred from local user control.
%
\begin{figure*}[t]
    \centering
    \includegraphics[width=1\textwidth]{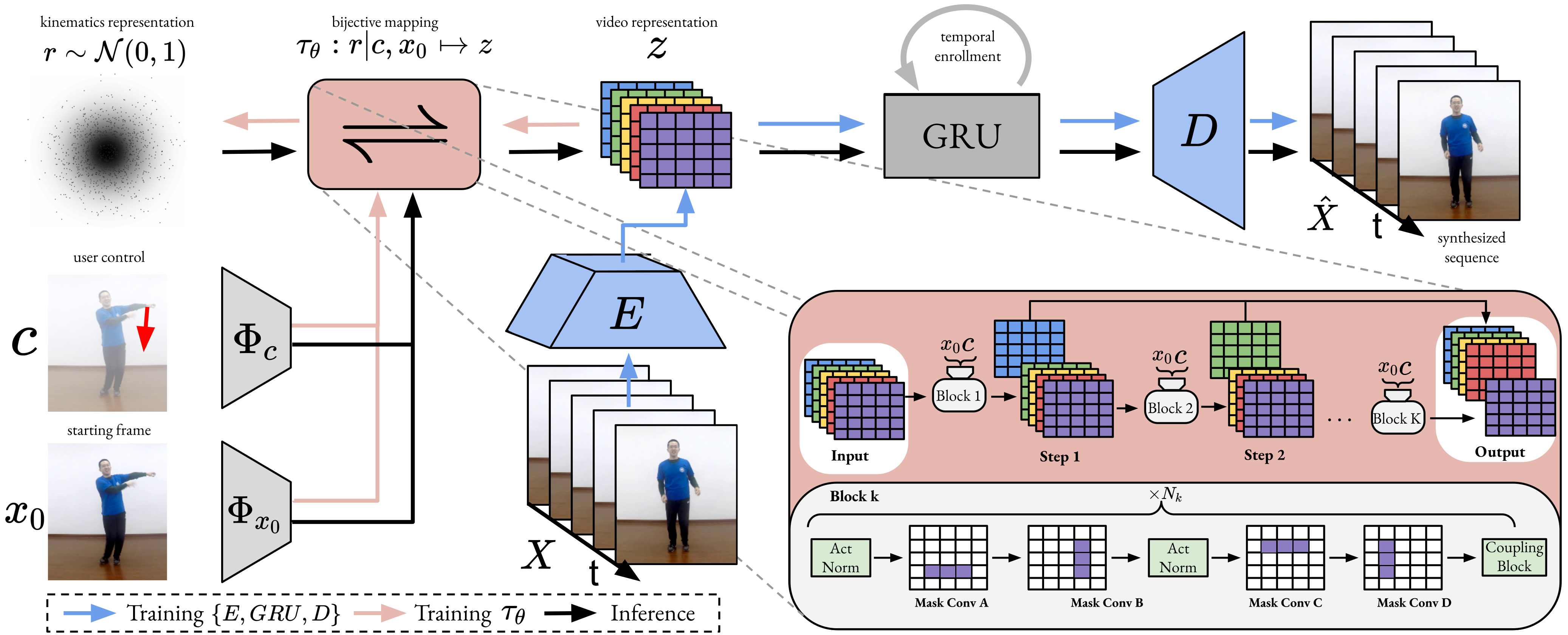}
    \caption{
    \textit{Overview of our proposed framework iPOKE for controlled video synthesis:} We apply a conditional bijective transformation $\tau_{\theta}$ to learn a residual kinematics representation $r$ capturing all video information not present in the user control $c$ defining intended local object motion in an image frame $x_0$ (orange path). To retain feasible computational complexity, we pre-train a video autoencoding framework $(E, GRU, D)$ (blue path) yielding a dedicated video representation $z$ as training input for $\tau_\theta$. Controlled video synthesis is achieved by sampling a residual $r$, thus defining plausible motion for the remaining object parts not directly affected by $c$, and generating video sequences $\hat{\boldsymbol{X}}$ from the resulting $z = \tau_{\theta}(r \vert x_0,c)$ using $GRU$ and $D$ (black path).}
    \label{fig:method}
    \vspace{-4mm}
\end{figure*}

\section{Related Work}
\textbf{Video Synthesis.} 
Video synthesis denotes the general task of generating novel video sequences. While some works solely focus on transferring a predefined holistic motion between objects~\cite{vid2vid} or interpolating motion between a starting and end frame~\cite{Niklaus_2018_CVPR,xue2019video,liu2019cyclicgen,Bao_2019_CVPR,Niklaus_2020_CVPR}, the most commonly addressed problem is video prediction. Given an initially observed video sequence, the goal is to infer a likely continuation into the future. To this end, proposed methods either generate a single, deterministic video sequence~\cite{2018epva,villegas2018decomposing,van2018relational,wu2020future,poke_blattmann_2021} or model the distribution over likely future sequences~\cite{cdna2016,svg,2018savp,sdcnet18,minderer2020unsupervised,vrnn-hier,cinn_dorkenwald_2021}. Moreover, the employed model architectures exhibit large divergence with latent RNN-based methods being the dominant modelling choice~\cite{minderer2020unsupervised,Franceschi2020}. However, also more complex models based on transformers~\cite{Weissenborn2020Scaling}, pixel-level autoregression~\cite{matthieu16,2018savp,LuHirSch17,svg,villegas2018decomposing,vrnn-hier}, factorization of dynamics and content~\cite{minderer2020unsupervised,Franceschi2020} and image warping using optical flow~\cite{vondrick5,voxel_flow,Gao_2019_ICCV} have been proposed. Despite these methods showing promising results, none of them is able to exercise control over the video generation process.
\\
\noindent \textbf{Controllable Video Synthesis.} 
Exercising user control over the video synthesis process requires a detailed understanding of the object kinematics and interplay of the object parts. To circumvent the difficult task of learning object kinematics directly from data, Davis et al.~\cite{davis2015} resort to fixed, linear mathematical models. Thus, they can only consider constraint oscillating motion around an object's rest state. In contrast, our model learns natural, unconstrained object kinematics from video, thus is also applicable to highly complex articulation such as those of humans.
Other works rely on a low-dimensional, parametric representation e.g. keypoints to transfer motion between videos~\cite{aberman2020unpaired,behavior_blattmann_2021} or to synthesize videos based on action labels~\cite{Yang_2018_ECCV}. Given such assumptions, these works cannot be universally applied to arbitrary object categories and allow only for coarse control compared to our fine-grained, local object manipulations.
By iteratively warping single images with local sets of estimated optical flow vectors, \cite{controllable_image} takes a first step towards sparsely controlled video generation for arbitrary object categories. However, due to the method's warping based nature, it is still not able to generate temporally coherent motion and requires optical flow guidance for each individual predicted image frame. To overcome such limitations, \cite{poke_blattmann_2021} introduces a hierarchical dynamics model, which can predict complex object dynamics controlled by a single optical flow vector in a given image, but does still not consider the natural motion ambiguity of the remaining, uncontrolled object parts. In contrast, our model learns a dedicated, stochastic kinematics representation modeling this incertitude of the object remainder and, thus, is capable of synthesizing locally controlled but also diverse object motion.\\
\noindent \textbf{Invertible Neural Networks.} 
Invertible neural networks (INNs) are learnable bijective functions often used to transform between two probability distributions, thus being a natural choice for addressing inverse problems~\cite{ardizzone2019analyzing}, introspecting and explaining neural network representations~\cite{esser2020disentangling,jacobsen2018irevnet} and domain transfer \cite{rombach2020network, rombach2_network}. Typically, INNs are realized as generative normalizing flows~\cite{rezende2016variational,lugmayr2020srflow,glow} which have recently also found application in image~\cite{glow, pumarola2020cflow} and video synthesis~\cite{kumar2020videoflow,cinn_dorkenwald_2021}. In this work, we use normalizing flows to learn the missing residual information, i.e. the latent object kinematics, not being determined by the by the sparse local control over part of the object motion.
\enlargethispage{\baselineskip}
\section{Approach}
Controlled video synthesis seeks to generate a plausible future video sequence $\boldsymbol{X} \in \mathbb{R}^{T \times H \times W \times 3}$ given an initial frame $x_0$ and a user-defined control $c$ that locally specifies part of the video dynamics,
\begin{equation}
(x_0, c) \mapsto \boldsymbol{X} = [x_1,\dots,x_T] \,.
\label{eq:det}
\end{equation}
Our goal is here to efficiently control video synthesis. Instead of having users tediously specify the dynamics at each pixel, e.g. by providing a dense vector field \cite{wu2020future}, $c$ should only be a very sparse signal. Thus, we assume to be provided only a local poke, the desired movement at one image location between start and end frame. The poke $c\in \mathbb{R}^4$ consequently comprises a shift, $c_{1:2}$, at a single pixel location, $c_{3:4}$, performed only by a simple mouse drag. 
Evidently, even densely defining the motion of every pixel between start and end frame does not fully define the object dynamics in between, even less so only a sparse 4D $c$ vector. 
Given this highly limited conditioning information, we model the distribution of all plausible future videos
\begin{equation}
\boldsymbol{X} \sim p(\boldsymbol{X}|x_0, c) \, ,
\label{eq:general_problem1}
\end{equation}
thus contrasting previous work, which only yields some arbitrary, uncontrolled realization~\cite{2018savp,LuHirSch17,svg,vrnn-hier}.
Our main challenge is then to model the \emph{object kinematics} which define how the movement of one part of an object affects the rest, thus yielding overall concerted object dynamics. As $\boldsymbol{X}$ is a random variable, the mapping in \eqref{eq:det} is actually non-unique. There is a lot of residual information $r$ beyond user control, which we need to turn \eqref{eq:det} into a unique one-to-one mapping
%
\begin{equation}
(x_0, c, r) \mapsto \boldsymbol{X} \,,
\label{eq:prob}
\end{equation}
where the residual $r$ would then capture object kinematics specifying the movement of the remaining object parts given the sparse local control $c$.
\subsection{Invertible Controlled Video Synthesis}
\label{sec:invertible_video_synthesis}
\noindent
Seeking to find the mapping \eqref{eq:prob} we naturally arrive at a problem of stochastic video prediction. So far, the dominant approach to such problems are conditional variational autoencoder (cVAE) based models~\cite{VAE, rezende2014stochastic, cvae}. cVAE employs strong regularization to remove the given conditioning from the remaining data variations, thus facing a \emph{trade-off} between synthesis quality and capturing all these variations~\cite{chen2016variational, zhao2017infovae}, in our case the diverse object kinematics $r$. 
\\
To avoid this, we use a conditional bijective, i.e. one-to-one, mapping $\tau$ between each residual $r$ and the corresponding video 
\begin{equation}
\boldsymbol{X} = \tau(r\vert x_0, c)
\label{eq:dyn_to_vid}
\end{equation}
so that all plausible $\boldsymbol{X}$ for a given conditioning can be synthesized. 
Moreover, the inverse mapping $\tau^{-1}$ allows to recover the residual kinematics for any $\boldsymbol{X}$,
\begin{equation}
r = \tau^{-1}(\boldsymbol{X}\vert x_0, c) \,,
\label{eq:vid_to_dyn}
\end{equation}
which then can be considered as a random variable $r~\sim~p(r|x_0, c)$, since $\tau^{-1}$ is unique and $\boldsymbol{X}$ is a random variable defined in \eqref{eq:general_problem1}.
To solve the conditional video synthesis task, we now show how to learn $\tau$ such that $r$ \emph{(i)} indeed contains all video information not present in $(x_0, c)$ and \emph{(ii)} follows a distribution which can be easily sampled from. 
\begin{figure}[t]
\includegraphics[width=1\linewidth]{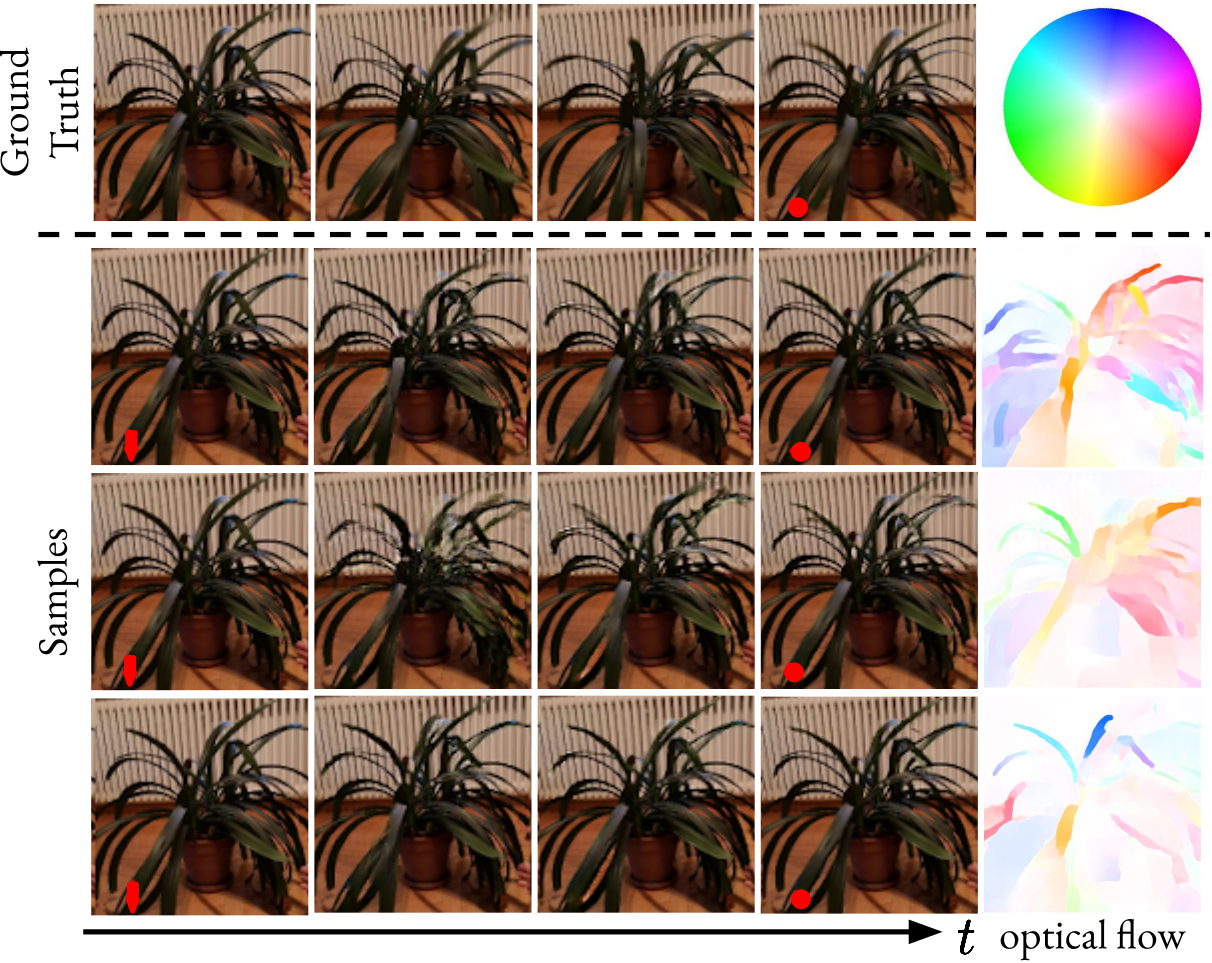}
\caption{Controlled stochastic video synthesis showing three video sequences for the same user control $c$ (red arrow) and randomly sampled kinematics $r$ on the PP dataset. Our model generates diverse, plausible object motion while accurately approaching the target location (red dot) for the controlled object part. Additionally, to ease comparison of the motion difference between samples, we show optical flow maps between the first and last frame of each sequence. Best viewed in video on our \href{https://bit.ly/3dJN4Lf}{project page}.}
\label{fig:plants}
\vspace{-6mm}
\end{figure}
\\
\noindent
\textbf{Learning the invertible mapping $\tau$.}
We equip $\tau$ with parameters $\theta$ which, by employing Eq.~\eqref{eq:vid_to_dyn}, can be learned from training videos $\boldsymbol{X}$. By the change-of-variables theorem for probability distributions, we have 
\begin{align}
p(\boldsymbol{X} \vert x_0, c) &= \frac{p(\tau_{\theta}(r \vert x_0, c)\vert x_0, c)}{\vert \det J_{\tau_{\theta}}(r \vert x_0, c) \vert} \notag \\ 
&=  p(\tau_{\theta}^{-1}(\boldsymbol{X} \vert x_0, c)\vert x_0, c    )  \cdot \vert \det J_{\tau_{\theta}^{-1}}(\boldsymbol{X} \vert x_0, c) \vert \; .
\label{eq:transformation}
\end{align}
Here, $J_{\tau_{\theta}}$ denotes the Jacobian of the transformation $\tau_{\theta}$ and $\vert \det [\cdot] \vert $ the absolute value of the determinant. Recall that we have to ensure to learn $\tau_{\theta}$ such that $r$ contains all video information \textit{not} present in $(x_0,c)$. 
Effectively, this requires learning $\tau_{\theta}$ such that $r$ is independent of $(x_0,c)$. This can be achieved by introducing some independent prior $q(r)$ and minimizing $\text{KL}[p(r \vert x_0, c) \Vert q(r)]$, which then constitutes an upper bound on the mutual information $\text{MI}\left[r, (x_0,c)\right]$~\cite{alemi2016,rombach2020network} as derived in Appendix~\ref{ap:independence} and thus, indeed forces the intended independence.
Moreover, by using Eq.~\eqref{eq:vid_to_dyn} and \eqref{eq:transformation}, we can express $\text{KL}[p(r \vert x_0, c) \Vert q(r)]$ as a function of the training data $\boldsymbol{X}$ what facilitates learning of $\tau_{\theta}$ by minimizing
\begin{align}
\text{KL}[p(r \vert x_0, c) \Vert q(r)] 
     \propto & ~\mathbb{E}_{\boldsymbol{X}}\Big[-\log \left( q\left(\tau_{\theta}^{-1}(\boldsymbol{X} \vert x_0,c)\right)\right) \notag \\ 
    \label{eq:loss_raw}&- \log  \vert \det J_{\tau_{\theta}^{-1}}(\boldsymbol{X} \vert x_0, c) \vert \Big]\, .
\end{align}
By selecting $q(r) = \mathcal{N}(r \vert 0, \mathbf{I})$~\cite{VAE,yan2016cVAE} and inserting this into Eq.~\eqref{eq:loss_raw} we arrive at the simple objective function
\begin{equation}
\begin{split}
 \min_{\theta} \mathcal{L}(\tau_{\theta},\boldsymbol{X},x_0,c) =& \; \mathbb{E}_{\boldsymbol{X}, x_0, c} [ \Vert \tau_{\theta}^{-1}(\boldsymbol{X} \vert x_0, c) \Vert_2^2 \\
& -  \log \vert \det J_{\tau_{\theta}^{-1}}(\boldsymbol{X} \vert x_0, c) \vert] \:.
\end{split}
\label{eq:obj_fn}    
\end{equation}
A detailed derivation can be found in the Appendix~\ref{ap:loss}.
Note, that optimizing Eq.~\eqref{eq:obj_fn} \emph{simultaneously} ensures \emph{(i)} independence of $r$ and $(x_0,c)$ \emph{and} \emph{(ii)} yields a generative probabilistic model as we can easily draw samples from $q(r)$ and use the conditional mapping~\eqref{eq:dyn_to_vid} to obtain synthesized videos. Thus, our model is capable to synthesize videos in a controlled but nonetheless stochastic manner without facing the trade-off encountered in cVAE.
\begin{figure}[t]
\includegraphics[width=0.97\linewidth]{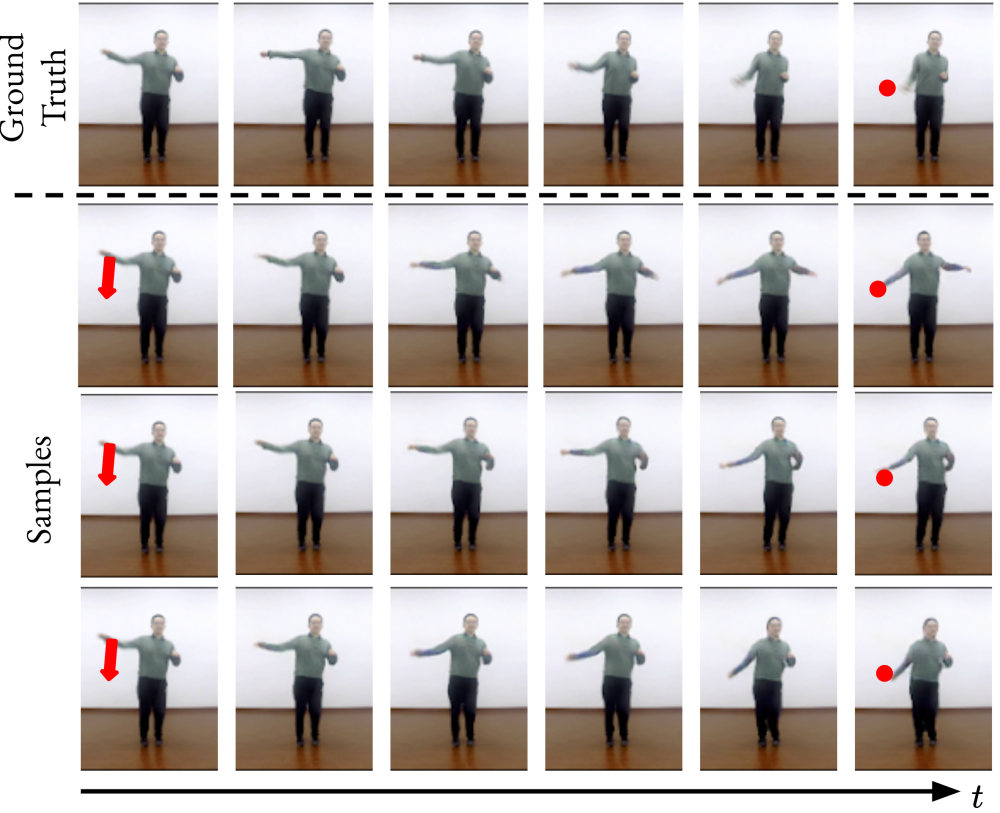}
 \caption{Controlled stochastic video synthesis showing three video sequences for the same user control $c$ (red arrow) and randomly sampled kinematics $r$ on the iPER dataset. Our model generates diverse, plausible object motion while accurately approaching the target location (red dot) for the controlled body part.  Best viewed in video on our \href{https://bit.ly/3dJN4Lf}{project page}.}
\label{fig:iPER}
\vspace{-6mm}
\end{figure}
\subsection{Architecture for Tractably Learning $\tau_{\theta}$}
\label{sec:model_arch}
\enlargethispage{\baselineskip}
\noindent
To realize the conditionally bijective nature of our mapping $\tau_{\theta}$, we implement it as a conditional invertible neural network (cINN)~\cite{papamakarios2019normalizing, dinh2017density, dinh2015nice, rombach2020network, cinn_dorkenwald_2021}, which requires equal dimensionality of the transformed random variables. Thus, $\boldsymbol{X}$ would demand $r$ to be very high dimensional, entailing infeasible computational complexity. As a remedy, we replace $\boldsymbol{X}$ with a compact, information-preserving video encoding $z \in \mathbb{R}^{h \times w \times d}$, with $ h \cdot w \cdot d  \ll H \cdot W \cdot 3 \cdot T$, learned by a standard sequence autoencoding framework~\cite{VAE} consisting of a 3D-ResNet~\cite{he2015ResNet} encoder $E$, a GRU~\cite{gru} for temporal enrollment in the latent space, and an image decoder $G$ to obtain video predictions $\hat{\boldsymbol{X}}$. Prior to learning $\tau_{\theta}$, we train this model to reconstruct training videos $\boldsymbol{X}$ by using a respective loss $\mathcal{L}_{rec}$ and additionally add static and temporal discriminators~\cite{DVDGAN, vid2vid}, $\mathcal{D}_{S}$ and $\mathcal{D}_{T}$, to increase visual and temporal coherence of $\hat{\boldsymbol{X}}$, thus resulting in the objective
\begin{equation}
\label{eq:ae_loss}
\mathcal{L}_{ae} = \mathcal{L}_{rec} + \mathcal{L}_{\mathcal{D}_{S}} + \mathcal{L}_{\mathcal{D}_{T}} \, .
\end{equation}
Detailed information on implementation and training can be found in the Appendix~\ref{ap:ae}. Afterwards we can learn $\tau_{\theta}$ from the compact  latent video encodings $z = E(\boldsymbol{X})$ instead of high-dimensional videos $\boldsymbol{X}$.\\
\noindent So far, cINNs operating on latent representations have been implemented as a sequence of fully connected layers~\cite{jacobsen2018irevnet,rombach2020network,rombach2_network}, thus discarding the spatial information naturally constituting visual data. However, since the conditioning $c$ describes a spatial shift of a single pixel, such architectures are not able to effectively leverage this information. To this end, we use the poke $c$ to define a two-channel map $C \in \mathbb{R}^{H \times W \times 2}$ with $C_{c_3,c_4,1:2} = c_{1:2}$ and zeros elsewhere, and instead design a fully convolutional  cINN, such that the crucial spatial information about the control location can be incorporated as best as possible. More specifically, our architecture comprises $K$ subsequently arranged cINN sub-blocks. By directly forwarding a portion $\frac{d}{K}$ of the output of each block to the final representation $r$, we reduce memory requirements and avoid vanishing gradients for large $K$~\cite{dinh2017density,macow_2019}. Within the $k$-th block, we apply a series of $N_k$ masked convolutions~\cite{macow_2019}, which have been shown to obtain improved expressivity compared to standard flow architectures such as coupling layers~\cite{dinh2015nice,dinh2017density,glow}. Finally, the conditioning information $(x_0,c)$ is separately processed by two dedicated encoding networks $\Phi_{x_0}$ and $\Phi_c$, yielding representations of the same spatial size than the flow input, to which they are concatenated before each masked convolution. We visualize the architecture and training in Fig.~\ref{fig:method} and provide further details in Appendix~\ref{ap:cinn}.
\begin{figure}[t]
\includegraphics[width=1\linewidth]{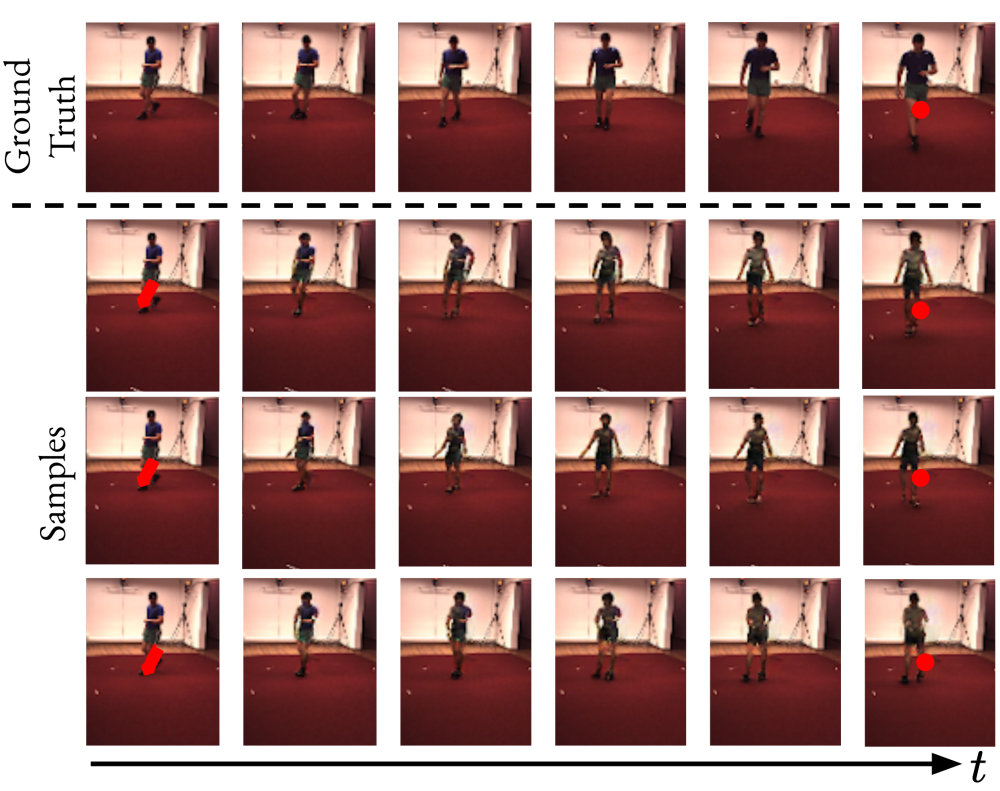}
\caption{Controlled stochastic video synthesis showing three video sequences for the same user control $c$ (red arrow) and randomly sampled kinematics $r$ on the Human3.6m dataset. Our model generates diverse, plausible object motion while accurately approaching the target location (red dot) for the controlled body part. Best viewed in video on our \href{https://bit.ly/3dJN4Lf}{project page}.}
\label{fig:h36m}
\vspace{-5mm}
\end{figure}
\subsection{Automatic Simulation of User Control}
\label{sec:poke_sim}
\enlargethispage{\baselineskip}
\vspace{-2mm}
\noindent
Training our model for controlled video synthesis relies on user controls $c$ and corresponding video sequences $\boldsymbol{X}$ depicting natural object responses to be available. Providing sufficient amounts of such training data for every targeted object category is tedious and costly. Instead, we employ an efficient self-supervised strategy to artificially generate such interactions directly from the observed motion of a collection of cheaply available training videos $\boldsymbol{X}$. To this end, we extract dense optical flow maps~\cite{flownet2} $F \in \mathbb{R}^{H \times W \times 2}$ between the start and end frames, $x_0$ and $x_T$, of $\boldsymbol{X}$ whose individual shift vectors can be interpreted as sparse pixel displacements $c = \{(F_{l_{n,1},l_{n,2},1},F_{l_{n,1},l_{n,2},2})\}_{n=1}^{N_c}$. During training, we randomly sample such simulated pokes at positions $l_n$ which exhibit sufficiently large motion that reliably corresponds to the foreground object. Contrasting~\cite{poke_blattmann_2021}, which use a similar strategy, but restrict the user control to be defined by only a single poke, we allow a user to control the degree of freedom of the object articulation by training our model on up to $5$ simultaneous interactions $c$, i.e. on a variable number $N_c \in [1,5]$ of local pokes. Note that for inferring user controls after training we do not require optical flow estimates, but use simple mouse drags instead.
\begin{figure}[t]
\includegraphics[width=1\linewidth]{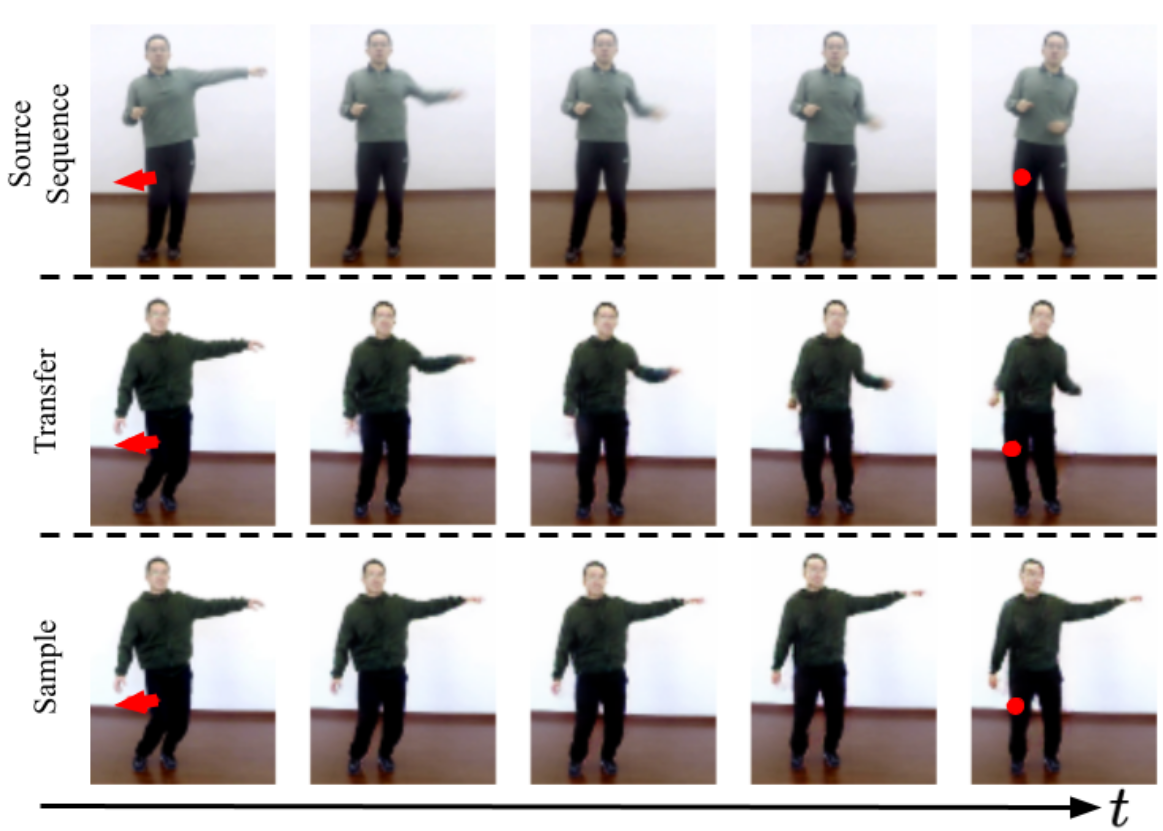}
\caption{\textit{Motion Transfer on iPER:} We extract the residual kinematics from a ground truth sequence (top row) and use it together with the corresponding control $c$ (red arrow) to animate an image $x_t$ showing similar initial object posture (second row). We also visualize a random sample from $q(r)$ for the same $(x_t,c)$ (bottom row), indicating that the residual kinematics representation solely contains motion information not present in $(x_t,c)$  (for a detailed description cf. Sec.~\ref{sec:eval_qual}). Best viewed in video on our \href{https://bit.ly/3dJN4Lf}{project page}.}
\label{fig:transfer}
\vspace{-3mm}
\end{figure}
%
\section{Experiments}
\vspace{-2mm}
Subsequently, we evaluate our model for controlled stochastic video synthesis on four video datasets showing diverse and articulated object categories of humans and plants. Implementation details and video material can be found in the Appendices~\ref{ap:impl_baselines} and \ref{ap:eval}, and on our \href{https://bit.ly/3dJN4Lf}{project page}.
\begin{figure*}[t]
    \centering
    \includegraphics[width=\textwidth]{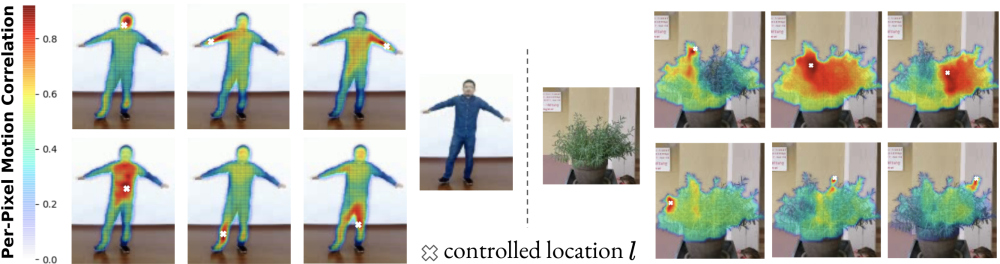}
    \caption{\textit{Understanding object kinematics:} By sampling 1000 random control inputs at location $l=c_{3:4}$ for a fixed image $x_0$ we obtain varying video sequences, from which we compute motion correlations for $l$ with all remaining pixels. By mapping these correlations to the pixel space, we visualize the interplay correlation of distinct object parts, thus yielding insights about the learned kinematics.}
    \label{fig:body_parts}
    \vspace{-1.5em}
\end{figure*}
\subsection{Datasets}
\enlargethispage{\baselineskip}
\vspace{-2mm}
\noindent
We evaluate our approach to understand and synthesize object dynamics on the following four datasets:
\\
\textbf{Poking-Plants (PP)}~\cite{poke_blattmann_2021} consists of 27 videos of 13 different types of pot plants. To learn a single kinematics model for all plants is notably challenging given the large variance in shape and texture of the plants. Overall, PP contains of 43k frames, from which a fifth is used as a test set and the remainder as training data. 
\\
\textbf{iPER} \cite{iper} contains of 30 humans with diverse styles performing various simple and complex movements. We follow the official train/test split which results in a training set of 180k frames and a test set of 49k frames. 
\\
\textbf{Tai-Chi-HD} \cite{first_order_2019} is a collection of 280 in-the-wild Tai-Chi videos from Youtube. We follow previous work~\cite{first_order_2019} and use 252 videos for training and 28 videos for testing. Given the large variance in background and camera movement, this dataset tests the real-world applicability of our model. Since the motion between subsequent frames is often small, we skip every other frame.
\\
\textbf{Human3.6m \cite{h36m}} is a large scale human motion dataset with video sequences of 7 human actors performing 17 different actions. We follow previous work~\cite{2018epva,minderer2020unsupervised,Franceschi2020} by centercropping and downsampling the videos to 6.25 Hz and by using actors S1,S5,S6,S7 and S8 (600 videos) for training  and actors S9 and S11 (239 videos) for testing.
\subsection{Qualitative Evaluation}
\label{sec:eval_qual}
\enlargethispage{\baselineskip}
\noindent
\textbf{Controlled Stochastic Video Synthesis.}
In Fig.~\ref{fig:plants}, \ref{fig:iPER} and \ref{fig:h36m} we show examples for controlled stochastic video synthesis generated by our proposed model on the PP, iPER and Human3.6m datasets. For each dataset, we show the ground-truth frames following a fixed given image $x_0$, as well as three synthesized examples generated from a fixed user control $c$ (red arrows) and randomly sampled kinematics realizations $r~\sim q(r)$. Examples for the Tai-Chi dataset can be found in the supplemental, where we also show additional synthesized videos based on control inputs from real human users and demonstrate our model to also plausibly react to different pokes at the same location. The individual videos are discussed in the Appendices~\ref{ap:add_vis} and \ref{ap:human}.
\\
\noindent
\textbf{Transfer of Kinematics.}
Besides sampling plausible object kinematics, we can also apply our model to transfer the kinematics inferred from a source sequence $\boldsymbol{X}_s = [x_{s,0}, \dots, x_{s,T}]$ to a novel object instance. To this end, we extract the corresponding residual kinematics $r_s = \tau_{\theta}^{-1}(\boldsymbol{X}_s \vert x_{s,0}, c)$ for a user control $c$ simulated based on $\boldsymbol{X}_s$ and use Eq.~\eqref{eq:dyn_to_vid} to animate a target image $x_t$ showing another object instance than $x_{s,0}$ with similar articulation. The resulting successfully transferred motion sequence $\hat{\boldsymbol{X}}_t = \tau_{\theta}(r_s \vert x_t, c)$ is shown in Fig.~\ref{fig:transfer} (second row) and compared to $\boldsymbol{X}_s$ (top row). It can be seen, that the motion contained in $\boldsymbol{X}_{s}$ is transferred to $\hat{\boldsymbol{X}}_t$ but \textit{not} the object appearance shown in $x_{0,s}$, indicating that our model indeed has learned a residual representation $r$ solely containing kinematics. Additionally, we visualize a synthesized video sequence based on a random sample $r \sim q(r)$ of residual kinematics for the same conditioning $(x_t,c)$ (bottom row), showing substantially different object motion except for the controlled body part and thereby providing evidence that $r$ is also independent of the user-control $c$. More results of kinematics transfer can be found in Appendix~\ref{ap:transfer}.
%
\begin{table*}[t]
	\centering
    \resizebox{.9\textwidth}{!}{
    \begin{tabular}{l|| ccc | ccc | ccc | ccc}
    \hline
     \multirow{2}{*}{Method} &
     \multicolumn{3}{c|}{\textbf{PP} \cite{poke_blattmann_2021}} & 
     \multicolumn{3}{c|}{\textbf{iPER} \cite{iper}} & 
     \multicolumn{3}{c|}{\textbf{Tai-Chi} \cite{first_order_2019}} & 
     \multicolumn{3}{c}{\textbf{Human3.6m} \cite{h36m}}  \\
      & {FVD $\downarrow$} & {LPIPS $\downarrow$} & {SSIM $\uparrow$} 
      & {FVD $\downarrow$} & {LPIPS $\downarrow$} & {SSIM $\uparrow$} 
      & {FVD $\downarrow$} & {LPIPS $\downarrow$} & {SSIM $\uparrow$}
      & {FVD $\downarrow$} & {LPIPS $\downarrow$} & {SSIM $\uparrow$}  \\
      \hline
      \hline
    Hao~\cite{controllable_image} & 361.51 & 0.16 & 0.72 & 235.08 &  0.11 & 0.88 & 341.79 & 0.12 & \textbf{0.78} & 259.92 & 0.10 & \textbf{0.93} \\
    Hao~\cite{controllable_image} w/ KP & -- & --  & --  & 141.07 & \textbf{0.04} & \textbf{0.93}  & --  & --  & --  & --  & --  & -- \\
    II2V~\cite{poke_blattmann_2021} & 174.18 & 0.10 & \textbf{0.78} & 220.34 & 0.07 & 0.89 & 167.94  & 0.12 & \textbf{0.78} & 129.62 & 0.08 & 0.91 \\
    \hline
    iPOKE (Ours) & \textbf{63.06} & \textbf{0.06} & 0.69 & \textbf{77.50} & 0.06 & 0.87 & \textbf{100.69} & \textbf{0.08} & 0.74 & \textbf{119.77}  & \textbf{0.06} & \textbf{0.93}\\
    \hline
  \end{tabular}
}
\caption{Comparison with recent methods for sparsely controlled video synthesis~\cite{controllable_image,poke_blattmann_2021}.}
    \label{tab:compare_ci}
    \vspace{-1.7em}
\end{table*}

\noindent\textbf{Understanding Object Kinematics.}
To demonstrate how well our model captures holistic object kinematics we analyze its understanding of the interplay of integral object parts. Therefore, we measure the pixel-wise correlations when applying $1000$ randomly sampled user controls $c$ at a fixed location $l$ of a fixed image $x_0$, i.e. varying only direction and magnitude of the shift vector. To measure the correlation in motion of all pixels with respect to the fixed control location (and thus the remaining object parts with the controlled part), we first compute optical flow maps between the start frame $x_0$ and the end frame $x_T$ of all resulting synthesized video sequences. Next, we compute the shift of the tracked pixel locations in $x_T$ with respect to the interaction location $l$, thus obtaining $1000$ $[\text{magnitude},\text{angle}]$ representations of the individual shifts. To measure the correlation of a pixel with $l$, we now compute the variance over these shifts. Fig.~\ref{fig:body_parts} illustrates the resulting correlation maps given different locations $l$ for both humans and plants. For humans, we obtain high correlations for pixels constituting a certain body parts and to parts which are naturally connected to $l$, showing our model correctly understands the body structure. For the plant, we see pulling at locations close to the trunk (top mid and right) intuitively affects large parts of the object. Interacting with individual small leaves mostly has only little effect on the remaining object, in contrast to the pixels representing the leave.
\subsection{Quantitative Evaluation}
\label{sec:eval_quan}
\enlargethispage{\baselineskip}
\noindent
As our proposed task of controlled \emph{and} stochastic video synthesis has been so far unattempted, we cannot directly compare iPOKE to previous work. To nevertheless quantitatively prove our model to reliably achieve this task, we separately compare against the current state of the art stochastic video prediction models~\cite{2018savp,vrnn-hier,Franceschi2020} and sparsely controlled video synthesis approaches~\cite{controllable_image, poke_blattmann_2021}. For all competitors we used the provided pretrained models, where available, or trained the models using official code.
%
\begin{figure}
\includegraphics[width=0.9\linewidth]{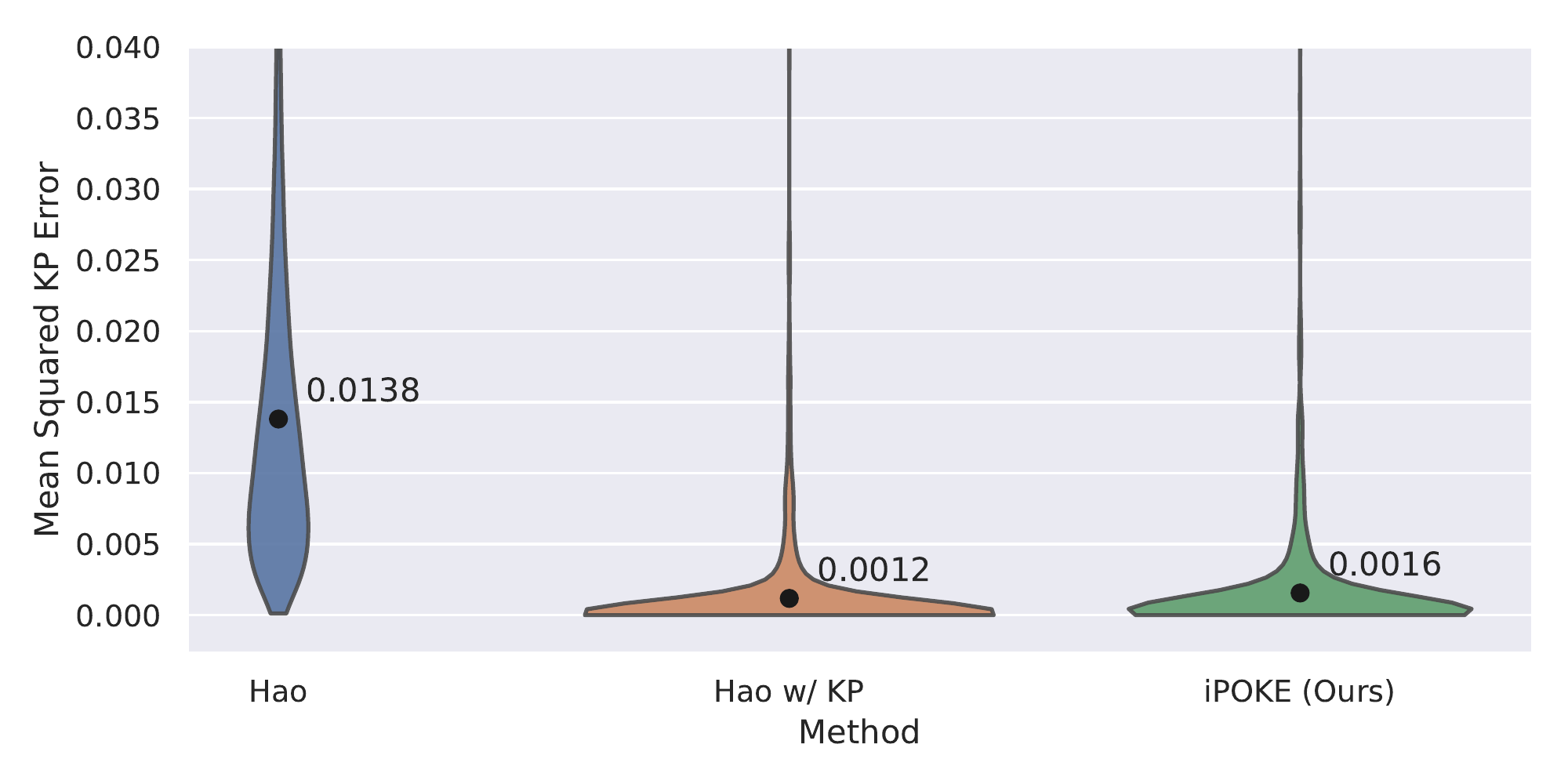}
   \caption{\textit{Control accuracy:} On the iPER dataset, we extract control signals $c$ based on ground truth keypoints and also estimate keypoints for the resulting synthesized videos. We only evaluate the errors with respect to those keypoints used to define $c$. The violins show the resulting MSE distributions. The numbers are the mean errors in keypoint space indicated by the black dots. Our model outperforms the baselines of Hao et al. by a large margin and even approaches their model which is trained on keypoints.}
\label{fig:kp_acc}
\vspace{-6mm}
\end{figure}

\begin{figure}[t]
\begin{minipage}[t]{0.49\linewidth}
\includegraphics[width=\linewidth]{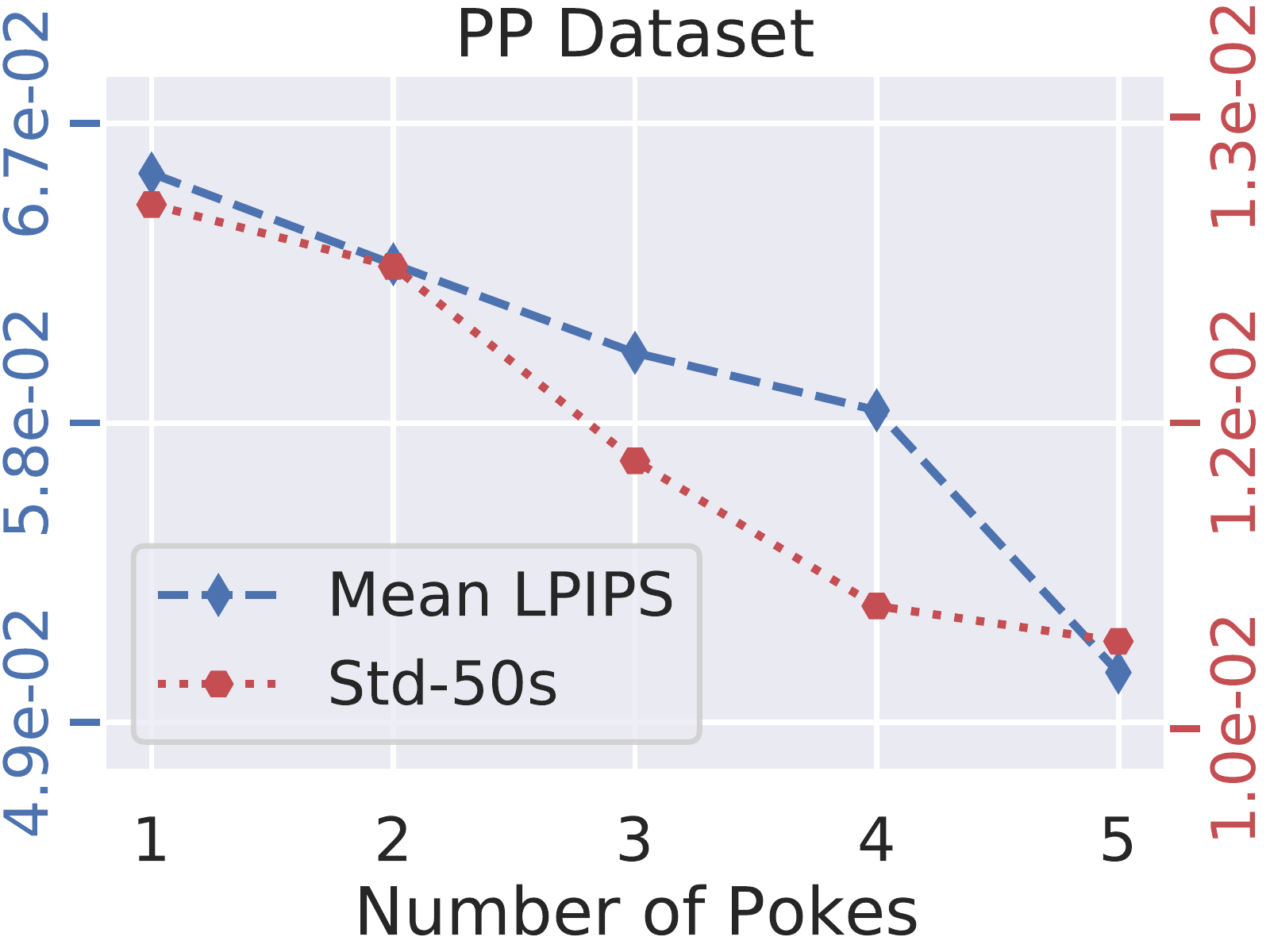}
\end{minipage}
\begin{minipage}[t]{0.49\linewidth}
    \includegraphics[width=\linewidth]{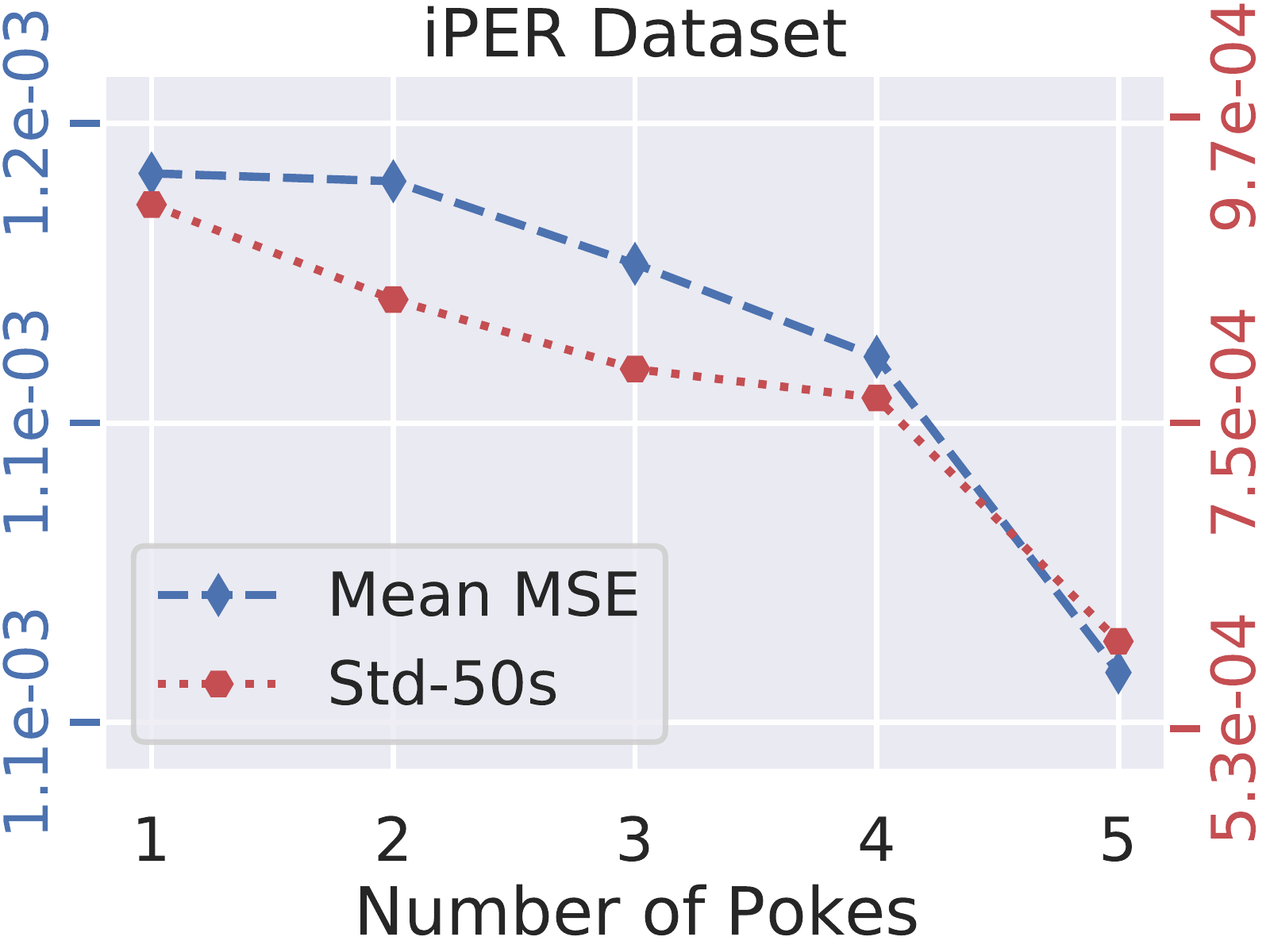}
\end{minipage}
   \caption{\textit{Controlling Future Ambiguity}: On the PP (left) and iPER (right) datasets our model reduces mean prediction errors (blue) and standard deviations of a sample of 50 residual samples given the same $(x_0,c)$ for an increased number of control inputs. Thus, our approach enables users to control future ambiguity by selecting the number of control inputs.}
\label{fig:kps_mean_var}
\vspace{-2mm}
\end{figure}

\noindent
\textbf{Evaluation Metrics.} 
\enlargethispage{\baselineskip}
\\
\textit{Motion Consistency.} We  evaluate  the  synthesis  quality  by using  the  Fr\'echet Video Distance~\cite{FVD}  (FVD, lower-is-better) which is responsive to visual as well as temporal coherence and uses an I3D  network~\cite{inception} trained on the Kinetics~\cite{Kinetics-original} dataset as backbone. Unterthiner et al.~\cite{FVD} showed that the metric correlates well with human judgement. The FVD-scores we report are obtained from video of length 10. \\
\noindent
\textit{Synthesis Quality.} Since we have no direct means to evaluate how well iPOKE models object kinematics, we compare its synthesized videos against the groundtruth using two commonly used framewise metrics, as producing uncorrect kinematics would lead to large errors between the individual generated and groundtruth frames. We average over time and over $5$ samples due to the stochastic nature of our model. As it has been shown to account for high- and low-frequency image differences and also to correlate well with human judgement, LPIPS~\cite{lpips} (lower-is-better) is the metric of choice for this task. Additionally, due to its wide application, we report framewise discrepancy as measured by SSIM~\cite{ssim}. However, as this metric compares image patches based on the L2 distance, it is known to be deceivable by blurry predictions.\\
\noindent
\textit{Motion Diversity}. Following previous work~\cite{2018savp, zhu2018multimodal} we evaluate the diversity by computing mutual distances between the individual frames of different video samples (while fixing the user control) using the LPIPS~\cite{lpips} metric. Moreover, we also directly evaluate the diversity in the pixel space using the MSE, thus measuring low-frequency image differences. 
\begin{table*}[t]
    \resizebox{\textwidth}{!}{
    \begin{tabular}{l|| ccc | ccc | ccc | ccc}
    \hline
     \multirow{2}{*}{Method} &
     \multicolumn{3}{c|}{\textbf{PP}} & 
     \multicolumn{3}{c|}{\textbf{iPER} \cite{iper}} & 
     \multicolumn{3}{c|}{\textbf{Tai-Chi} \cite{first_order_2019}} & 
     \multicolumn{3}{c}{\textbf{Human3.6m} \cite{h36m}}  \\
      & {FVD $\downarrow$} & {DIV MSE$^\ddagger \uparrow$} & {DIV LPIPS$^\ddagger \uparrow$} 
      & {FVD $\downarrow$} & {DIV MSE$^\ddagger \uparrow$} & {DIV LPIPS$^\ddagger \uparrow$} 
      & {FVD $\downarrow$} & {DIV MSE$^\ddagger \uparrow$} & {DIV LPIPS$^\ddagger \uparrow$}
      & {FVD $\downarrow$} & {DIV MSE$^\ddagger \uparrow$} & {DIV LPIPS$^\ddagger \uparrow$}  \\
      \hline
      \hline
    SAVP \cite{2018savp}$^\dagger$ & 92.2 & -- & -- & 92.8 & -- & -- & 236.8 & -- & -- & 131.7 & -- & --\\
    IVRNN \cite{vrnn-hier} & 128.3 & 2.52 & 8.23 & 126.0 & 37.66 & 93.26  & 150.2 & 0.34 & 1.65 & 238.6 & 46.45 & 106.71\\
    SRVP \cite{Franceschi2020} & 171.9 & 110.37 & 225.77 & 274.2 & 53.94 & 164.46 &  268.9 & 30.2 & 16.00 & 140.1 & 93.61 & 224.07\\
    \hline
    iPOKE (Ours) & \textbf{56.59} & \textbf{133.37} & \textbf{275.04} & \textbf{81.49} & \textbf{98.95} & \textbf{201.58} & \textbf{96.09} & \textbf{69.96} & \textbf{126.76} & \textbf{111.55}  & \textbf{124.25} & \textbf{309.06}\\
    \hline
  \end{tabular}
}
\caption{Comparison with recent state-of-the-art in stochastic video prediction. As our model does not face trade-off between variability and synthesis quality, we obtain significantly better motion video quality and diversity scores for all considered datasets. $^\dagger$: SAVP faced mode collapse due to training instabilities caused by the two involved discriminator networks. As a consequence their model generates entirely equal outputs when sampling. Therefore, we are unable to report diversity scores for this baseline. $^\ddagger$: Reported numbers multiplied with $1e4$.}
\label{tab:comp_vp}
\vspace{-4mm}
\end{table*}
\\
\noindent
\textbf{Controllable Video Synthesis.}
We compare our model with the considered methods for sparsely controlled video synthesis~\cite{controllable_image,poke_blattmann_2021} on all considered datasets using LPIPS~\cite{lpips}, SSIM~\cite{ssim} and FVD~\cite{FVD} on images of resolution of $128\times 128$. Note that both competing baselines are limited in that they provide no means to stochastically model the inherent ambiguity of the non-controlled object parts. Additionally,~\cite{controllable_image} lacks a dedicated dynamics model, as this method is based on a warping technique, which we describe in Appendix~\ref{ap:impl_baselines}, and requires more than one control inputs to reliably generate complex object articulation. Due to these limitations, our model exhibits significantly better temporal and visual consistency as indicated by the large gaps in FVD and LPIPS scores in Tab.~\ref{tab:compare_ci}. To provide a stronger baseline, we also train and evaluate the model of Hao et al. with input trajectories based on groundtruth keypoints (Hao w/KP) which are readily available for the iPER dataset and much more reliable than those based on estimated optical flow. Despite this advantage, we also outperform this baseline in FVD and generate similarly sharp image frames as indicated by comparable LPIPS scores.\\
Next, we use the displacements between the groundtruth keypoints of the test sequences to construct targeted user controls for each individual part of the human body. By using these manipulations as test-time inputs and estimating keypoints~\cite{hr_net} for the resulting generated videos, we assess the targeted control accuracy by measuring the Mean Squared Error (MSE) only between those estimated and groundtruth keypoints which correspond to the poked body parts. Fig.~\ref{fig:kp_acc} shows the resulting error distributions and means (black dots) showing that we significantly outperform Hao et al.~\cite{controllable_image} and achieve similar performance to their keypoint-based version. Thus our model allows for accurate control of body parts which are correctly moved to the intended target locations.  
\\
\noindent \textbf{Stochastic Video Synthesis.} 
\enlargethispage{\baselineskip}
To evaluate the visual quality and the diversity of generated videos we compare against recent state of the art methods for stochastic video synthesis (SVS)~\cite{2018savp,minderer2020unsupervised,Franceschi2020}, each of them based on variational autoencoder (VAE).  We adopt the SVS evaluation protocol and generate videos of spatial size $64 \times 64$. Tab.~\ref{tab:comp_vp} summarizes the comparison in video quality (measured in FVD score) and sample diversity (measured using LPIPS and pixel-space MSE). All SVS methods are conditioned on two image frames directly preceding the predicted sequence if not stated otherwise. Details for training and evaluation protocols can be found in the Appendices~\ref{ap:impl_baselines} and \ref{ap:eval}. Our method outperforms all competing approaches by large margins in both video quality and diversity. Moreover, Tab.~\ref{tab:comp_vp} reveals that competing methods which achieve comparable FVD scores to ours, i.e. video synthesis of similar visual quality, fail in generating diverse samples and vice versa. We attribute the limited performance of these models to the discussed trade-off in synthesis quality and capturing data variations of VAE-based approaches (cf. Sec.~\ref{sec:invertible_video_synthesis}).
\\
\noindent
\textbf{Controlling Future Ambiguity.}
We now assess the ability of our model to control the degree of freedom in stochastic object articulation by varying the number of local pokes. Intuitively, an increasing number of user controls should result in more accurate predictions and lowered between-sample-variance due to the reduction in future ambiguity. We evaluate the amount of uncertainty in predictions by comparing average reconstruction scores of a fixed number of samples from $q(r)$ for increasing numbers of user controls. More specifically, we report the mean prediction error and standard deviation of 50 samples (Std-50s) for each of 1000 input images and pokes. On the iPER dataset this is done by measuring MSE between estimated~\cite{hr_net} and groundtruth keypoints. For PP dataset we resort to the LPIPS metrics as keypoints are not available. The resulting curves are depicted in Fig.~\ref{fig:kps_mean_var}. As expected, the decreasing prediction errors and between-sample-variances indicate that our model leverages the additional future information provided by an increased number of pokes. Thus, our model not only generates diverse predictions but also provides means to control their uncertainty by choosing an appropriate number of input pokes.   
\\
\textbf{Ablation Study.}
\enlargethispage{\baselineskip}
As the competing VAE-grounded baselines for SVS are all conditioned on observed motion in form of observed past frames rather than dedicated, local user control, we further compare our model method to a cVAE-baseline (Ours cVAE) for locally controlled video synthesis. Thus, we use the exact architecture of our video-autoencoding framework (cf. Sec.~\ref{sec:model_arch}) except for our latent cINN model. To enable sampling, we realize the latent video representation $z$ as a gaussian distribution and regularize it towards a standard normal prior. The encodings obtained from the control $c$ and source image $x_0$ are concatenated with $z$ and constitute the hidden state for the latent GRU. A detailed architecture and training description of the baseline is contained in the Appendix~\ref{ap:impl_baselines}. Thus, this baseline is the exact variational counterpart of our model. We conduct ablation experiments on all the considered object categories, using the PP and Human3.6m datasets. Tab.~\ref{tab:ablation}, which summarizes the results, again indicates the improved performance of our invertible model compared to variational approaches. 
\begin{table}[t]
    \resizebox{0.5\textwidth}{!}{
    \begin{tabular}{l|| ccc | ccc}
    \hline
     \multirow{2}{*}{Method} &
     \multicolumn{3}{c|}{\textbf{PP}} & 
      \multicolumn{3}{c}{\textbf{Human3.6m} \cite{h36m}} \\
      & {FVD $\downarrow$} & {DIV MSE $\uparrow$} & {DIV LPIPS $\uparrow$} 
      & {FVD $\downarrow$} & {DIV MSE $\uparrow$} & {DIV LPIPS $\uparrow$}  \\
      \hline
      \hline
    Ours cVAE & 70.9 & 3.37 & 7.59 & 269.6 & 83.17 & 210.39\\
    iPOKE (Ours) & \textbf{56.59} & \textbf{133.37} & \textbf{275.04} &\textbf{111.55}  & \textbf{124.25} & \textbf{309.06}\\
    \hline
  \end{tabular}
}
  \caption{\textit{Ablation}. Comparison with a cVAE-counterpart to our cINN-based model for controlled video synthesis, indicating its superior performance due to variability vs. quality trade-off in cVAE.}
  \label{tab:ablation}
  \vspace{-6mm}
\end{table}
\section{Conclusion}
\enlargethispage{\baselineskip}
\vspace{-2mm}
We presented a novel model for controlling and synthesizing object kinematics of arbitrary object categories by locally manipulating object articulation using simple mouse drags. Our model is based on an invertible mapping between the generated video sequences and a dedicated kinematics representation learned from training videos only. To account for the ambiguity in the global object articulation given a local shift determining the motion of only an object part, learning is based on a probabilistic formulation, thus allowing us to sample and synthesize diverse kinematic realizations.
\section*{Acknowledgements}
\enlargethispage{\baselineskip}
\vspace{-2mm}
This research is funded in part by the German Federal Ministry for Economic Affairs and Energy within the project “KI-Absicherung – Safe AI for automated driving” and by the German Research Foundation (DFG) within~project~421703927.

\clearpage
\onecolumn
\appendix
\begin{center}
\LARGE\textbf{Appendix}\\
\end{center}

\section*{Preliminaries}
For all of our reported datasets in the main paper, we provide additional video material resulting in 14 videos in total. Analogous to the main paper, the direction of the control input is indicated by a red arrow starting at the poke location $l$ and the target location is marked by a red dot, if not stated otherwise. For some examples were either the arrow depicting the user input or the target location are small, we additionally overlay them with larger arrows and dots. We loop each video three times and replay it with three fps.
The file structure of the videos is as follows:
\begin{verbatim}
  04284-supp
        |
        +--A-Additional_Visualizations
            |
            +--A1-Controlled Stochastic 
                  Video Synthesis
                |
                +--A1.1-Samples_PP.mp4
                |   
                +--A1.2-Samples_Iper.mp4
                |
                +--A1.3-Samples_Human36m.mp4
                |
                +--A1.4-Samples_TaiChi.mp4
            |
            +--A2-Kinematics_Transfer
            |
            +--A3-Control_Sensitivity
            |
        +--B-Control Inputs 
             of Human Users
            |
            +--B1-Human_Control_PP.mp4
            |   
            +--B2-Human_Control_Iper.mp4
            |
            +--B3-GUI-Demo_1.mp4
            |
            +--B3-GUI-Demo_2.mp4
            |
        +--C-Qualitative Comparison
             with Baselines
            |
            +--C1-Controllable 
                  Video Synthesis
                |
                +--C1.1-Cmp_Hao_PP.mp4
                |
                +--C1.2-Cmp_Hao_Iper.mp4
                |
            +--C2-Stochastic 
                  Video Synthesis
                |
                +--C2.1-Cmp_IVRNN_PP.mp4
                |
                +--C2.2-Cmp_IVRNN_Iper.mp4
                |
\end{verbatim}
In Section~\ref{ap:add_vis}, we show and discuss qualitative video results of iPOKE on the four considered datasets for the tasks of controllable stochastic video synthesis (Section~\ref{ap:sampling}) and transfer of kinematics between two unrelated object instances (Section~\ref{ap:transfer}), which are the two main tasks considered in our paper. Our model iPOKE, however, also allows human users to animate still images by poking, as demonstrated by examples shown in Section \ref{ap:human}, which further contains examples of interactions of our model with human users via a graphical user interface. For the quantitative evaluations conducted in the main paper, we additionally provide qualitative comparisons in Section~\ref{ap:comp_qual}. A derivation of our objective function is given in Section~\ref{ap:deriv_obj}. Moreover, we also provide the implementation details of our model in Section~\ref{ap:impl_ipoke} and these of all considered baselines in Section~\ref{ap:impl_baselines}. Finally, describe the metrics used for evaluation in more detailed in Section~\ref{ap:eval}.

\section{Additional Visualizations}
\label{ap:add_vis}
Here we show and discuss additional video material showing the results of iPOKE on all considered datasets for our main task of controllable stochastic video synthesis. Moreover we also provide additional examples of successfully transferred kinematics from a source sequence to a given still image frame on the iPER~\cite{iper} dataset.

\subsection{Controllable Stochastic Video Synthesis}
\label{ap:sampling}
\noindent
For each individual dataset presented in the main paper we show a video containing results of iPOKE for multiple object instances. Within each video, the first column depicts the ground truth video sequences starting with the same image frame $x_0$ which is used as part of the conditioning for our model. The remaining columns contain five generated video sequences corresponding to different samples drawn from the residual kinematics prior $q(r)$ for the same conditioning $(x_0, c)$. The user control $c$ is visualized as a red arrow starting at the controlled location $l = c_{3:4}$ and pointing to the intended target location, which is depicted by a red dot and plotted after the sequence has ended. As the arrows are sometimes tiny, we additionally visualize larger arrows indicating the direction of the user control input in the first column as an overlay over the ground truth sequences. For generating the user control, we used the procedure explained in Section~3.3 in the main paper. Subsequently, we discuss each video in detail.\\
\noindent
\textbf{A1.1-PokingPlants.}
For the PokingPlants dataset we show the results of iPOKE for five plants with substantially varying shapes and appearances in the video sequence \textit{A1.1-Samples\_Plants.mp4}. Despite these large variances, iPOKE manages to generate appealing video sequences showing plausible reactions to the user control $c$. In all cases, the intended end position of the controlled object part is reached. The remainder of the object, however, shows unique, plausible motion for each individual sample from $q(r)$ for the same conditioning. In summary, for the PokingPlants dataset, iPOKE is capable of rendering realistic video sequences by locally controlling the motion of a given part of the shown object instances. Moreover, diverse, plausible motion is generated for the remaining object parts.\\
\noindent
\textbf{A1.2-iPER.}
In the video \textit{A1.2-Samples\_Iper.mp4} we visualize results for the test set of iPER~\cite{iper}, thus showing persons previously unseen by iPOKE. We clearly see that our model generalizes to unseen appearances. Moreover, iPOKE generates realistic video sequences showing the poked human body parts approaching their intended target location, while retaining significant motion variance for these parts not directly affected by the user control $c$.\\
\noindent
\textbf{A1.3-Human3.6m.}
To also evaluate the capabilities of iPOKE to generate human motion not contained in iPER, i.e. behavior such as walking and sitting, we additionally use the Human3.6m~\cite{h36m} dataset. The resulting synthesized output of our model can be seen in the video \textit{samples\_human36m.mp4}. Also in this case, iPOKE achieves to generate realistic looking video sequences showing such complex human motion for control inputs ranging from very small (e.g. third row) to large (e.g. second and last rows) control inputs. Unfortunately, iPOKE changes the appearance of the depicted person if the generated motion is large. However, this is a common issue, when using the Human3.6m test set~\cite{minderer2020unsupervised,villegas2018decomposing,2018epva,Franceschi2020}, as the train set only contains five distinct appearances and, thus, generalization of appearance is very hard to achieve.\\ 
\noindent
\textbf{A1.4-Tai-Chi}
To further evaluate the capabilities of iPOKE to synthesize human motion for in-the-wild settings we use the Tai-Chi dataset~\cite{first_order_2019}. This is a very challenging dataset due to diverse video backgrounds, which also contains motion. The human motion contained in the Tai-Chi, however, is not as large and diverse as for the iPER and Human3.6m datasets. Nevertheless, our model is able to synthesize realistic looking sequences as visualized in \textit{A1.4-Samples\_Tai-Chi.mp4}.

\subsection{Kinematics Transfer}
\label{ap:transfer}
\noindent
In the video \textit{A2-Kinematics\_Transfer\_Iper.mp4} we show the ability of iPOKE to transfer kinematics from a source sequence (top row) to an image showing a person of arbitrary appearance exhibiting a similar starting posture (mid row). For comparison, we synthesized sequences in which we induce random object kinematics by sampling from our residual kinematics prior $q(r)$ (bottom row). For a further description on transfer of kinematics, cf. the respective paragraph in Section~4.2 in the main paper. The implications arising from observing these video examples are twofold: Firstly we can see that our model, due to its conditionally invertible nature, is capable of transferring kinematics from a given sequence to a still image in a fine-grained manner, as subtle details of the motion shown in the source sequence are preserved. Secondly, these experiments show that our residual kinematics representation $r$ is indeed independent of the conditioning factors $(x_0,c)$ as \textit{i)} our model only transfers kinematics from the source sequence to the target image and \emph{not} the appearance of the depicted person and \textit{ii)} the sequences obtained from drawing random samples from $q(r)$ exhibit significantly different motion compared to the corresponding transferred sequences \emph{except} for the controlled location, which is correctly approaching its target location in both cases. Thus, these two factors, are not influenced by the residual kinematics representation but only the kinematics of the body parts not directly influenced by the controlled location, due to this independence. This empirical evidence of the independence of $r$ from $(x_0,c)$ is further backed by theoretical arguments which are discussed in Section~\ref{ap:deriv_obj}. 

\subsection{Control Sensitivity}
\label{ap:control_sensititvity}
\noindent
In this paragraph, we directly verify that iPOKE also generates plausible object responses to control inputs $c$ at the same pixel location $c_{3:4}$ with different shift vectors $c_{1:2}$ for a given initial image frame $x_0$. To this end, we sample the initial frame $x_0$ of a sequence $\boldsymbol{X}$ from the iPER~\cite{iper} test set and simulate a vector $c$ based on the procedure introduced in Sec.~\ref{sec:poke_sim} from which we only take the respective interaction location $c^{*}_{3:4}$. To obtain different shift vectors at this location we randomly sample $L$ directions from a uniform distribution $\mathcal{U}(0,2\pi)$ and magnitudes from $\mathcal{U}(\mu_{\vert F \vert}, \max_{\vert F \vert})$, where $\mu_{\vert F \vert}$ and $\max_{\vert F \vert}$ denote the mean and maximum optical flow magnitudes in the optical flow map $F$ between $x_0$ and the end frame $x_T$ of $\boldsymbol{X}$.
Thus, we obtain $L$ different control inputs $\{(c^l, x_0) : c^{l}_{3:4} = c^{*}_{3:4}\}_{l=1}^{L}$ at the same location for a given $x_0$. Drawing different residual samples $\{r_l\}_{l=1}^{L}$ and applying the forward transformation results in video representations $z_l = \tau_{\theta}(r_l \vert x_0, c_l)$ which can afterwards be decoded to object responses by our autoencoding framework (cf. Sec.~\ref{sec:model_arch} and \ref{ap:ae}). Each row of the video \textit{A2-Control\_Sensitivity\_Iper.mp4} shows such generated object responses of iPOKE for $L=4$ different shift vectors for the same static input image $x_0$ and compares these to the ground truth video starting with $x_0$. From observing these results, we see that the the poked body part always approaches its intended target location which is defined by the respective shift vector $c^{l}_{1:2}$, thus indicating the sensitivity of our model with respect to the control input.

\section{Control Inputs of Human Users}
\label{ap:human}
So far, we have only considered simulated control inputs as described in Section~3.3 in the main paper. However, as our goal is not only to provide a model for controlled stochastic video synthesis, but also to enable human users to directly animate a still image frame, we now additionally show the results of our model for inputs of human users for the object categories of plants on the PP dataset in Sec.~\ref{ap:human_plants} and humans on iPER in Sec.~\ref{ap:human_iper}. Each video contains three unique object instances shown in the rows of the respective video \textit{B1-Human\_Control\_PP.mp4} \textit{B2-Human\_Control\_Iper.mp4}. In each row, we show the ground truth sequence on the left, followed by an example using a simulated control input as described in Sec.~3.3 in the main paper. In the remaining three rows, we show examples of control inputs by human users. The results were obtained via our graphical user interface (GUI) which is presented in the videos \textit{GUI\_Demo\_[1,2].mp4}  in Section~\ref{sec:gui}. The GUI will be open sourced together with the project code upon publication.   

\subsection{PokingPlants}
\label{ap:human_plants}
\noindent
The three plants in the video \textit{B1-Human\_Control\_PP.mp4} again show that iPOKE is capable of generating plausible object kinematics as a reaction to the control. In particular, these examples highlight the accurate object reactions generated by our model even for fine-grained user control as, for instance, shown in the third column of the first row, where a control input with a small magnitude at an outer trunk results very subtle motion, leaving most of the controlled plant mainly unaffected. Large pokes, on the other hand, which are depicted in the remaining columns of the first row, result in large amounts of motion, thus corresponding to what we humans would expect from such a manipulation. Moreover, the results visualize the generalization abilities of iPOKE, as not all of these pokes from humans are within the data distribution.

\subsection{iPER}
\label{ap:human_iper}
\noindent
The video \textit{B2-Human\_Control\_Iper.mp4} contains similar results to these discussed in the last section for the iPER dataset. The synthesized videos are realistic looking for inputs defined by human users and the controlled body parts are accurately moved to the intended target location. Moreover it is shown that our model learns to separate the pixels comprising the object from those in the background, as emerging from observing the fourth column in the top row, where a part of the background is controlled. In this case, our model does not generate human kinematics, but renders a still video sequence, where the entire person remains unaffected. This is caused by by our training procedure where we explicitly generate control signals in the background to teach the iPOKE to ignore such inputs. For a detailed description, refer to Section~\ref{ap:further_details}.
\subsection{Graphical User Interface}
\label{sec:gui}
\noindent
Finally we recorded two videos (\textit{B3-GUI\_Demo\_[1,2].mp4}) showing iPOKE interactively generating kinematics as defined by humans using our graphical user interface. The videos were recorded without cuts and, thus, clearly indicate the robustness of iPOKE. Moreover it can be seen that our model is capable of processing multiple control inputs one after another without generating heavy artifacts or implausible kinematics, thus enabling an interactive experience for human users. We plan to make this GUI publicly available and will release it together with our code.

\section{Qualitative Comparison with Competing Methods}
\label{ap:comp_qual}
To provide visual evidence for the quantitative evaluation presented in the main paper, we now show visual comparisons to Hao et al.~\cite{controllable_image} for the task of controllable video synthesis and to IVRNN~\cite{vrnn-hier} for the task of stochastic video synthesis. The comparisons with these methods are conducted in the iPER and PokingPlants (PP) datasets.

\subsection{Controllable Video Synthesis}
\label{ap:control}
\noindent
Similar to Sec.~\ref{ap:sampling}, in \textit{C1.1-Comparison\_Hao\_PokingPlants.mp4} and \textit{C1.2-Comparison\_Hao\_Iper.mp4} we show video generations for iPOKE and compare them to the approach of Hao et al~\cite{controllable_image} trained on sparse sets of random optical flow vectors and tracked keypoint representations (cf. Sec.~4.3 in main paper for details.). We observe that iPOKE produces fluent and realistic object kinematics on both datasets. In contrast, Hao et al. fails to capture and generate the complex, structured motion depicted in the iPER dataset due to only warping individual frames. We attribute the lower FVD scores obtained by this method, compared to ours, to this. On the PP dataset, the competing approach produces blurry outputs, as successive warping cannot represent coherent, fine-grained motion due to noisy optical flow estimates.

\subsection{Stochastic Video Synthesis}
\label{ap:stochastic}
\noindent
In \textit{C2.1-Comparison\_IVRNN\_Plants.mp4} and \textit{C2.2-Comparison\_IVRNN\_Iper.mp4} we compare video sequences generated by our model against sequences generated by IVRNN~\cite{vrnn-hier}, a state-of-the-art approach for unconditional stochastic video synthesis. We chose this particular competitor as it was the best performing of the competing methods on these two datasets when considering video quality and diversity together (cf. Tab.~2 in the main paper). In our case, we generate diverse video sequences depicting different object kinematic realizations inferred by sampling from the residual representation $r \sim q(r)$ for different fixed user interactions $c$ on the iPER and PP dataset, thus following a similar setup as in Sec.~\ref{ap:sampling}. For IVRNN~\cite{vrnn-hier} we show different video generations for two conditioning frames preceding the sequence to be generated. Our videos clearly show that our model iPOKE successfully synthesizes plausible video sequences which closely follow the user interaction $c$ while, on the other hand, the remaining object parts exhibit a diverse range of motion when varying $r$. In contrast, the videos samples of IVRNN hardly show visible differences across different video samples. Moreover, IVRNN fails to capture fine-grained motion, e.g. of individual leaves, on the PP dataset, thus generating blurry video sequences. This can be explained by the trade-off between quality and diversity, which IVRNN, as a VAE-based method, suffers from. 

\section{Derivation and Implications of Learning Objective}
\label{ap:deriv_obj}

\subsection{Independence of $r$ and $(x_0,c)$}
\label{ap:independence}
\noindent
We intend to learn $\tau$ defined in Eq.~\eqref{eq:dyn_to_vid} such that the residual $r$ contains all information about $\boldsymbol{X}$ not present in $(x_0,c)$. A way to achieve this is to force independence between $r$ and the conditioning $(x_0,c)$. We now show that a minimizer of $\text{KL}[p(r \vert x_0, c) \Vert q(r)]$, where $q(r)$ is some prior distribution independent of $(x_0,c)$, forces this independence. By the definition of the Kullback-Leibler-Divergence, we have
\begin{align}
    \text{KL}[p(r \vert x_0, c) \Vert q(r)] &= \int_{r,x_0,c} p(r,x_0,c) \log \left( \frac{p(r \vert x_0,c)}{q(r)} \right) \notag \\
                                            \label{eq:kl}&= \int_{r,x_0,c} p(r,x_0,c) \log \left(p(r \vert x_0,c) \right) - \int_{r}p(r)\log \left(q(r) \right) \; .
\end{align}
Following the argument of~\cite{rombach2_network,alemi2016} and using the fact that the Kullback-Leibler-Divergence is always positive, we have
\begin{align}
\label{eq:kl_pos} 
 \text{KL}[p(r) \Vert q(r)] &\geq 0 \Rightarrow \int_{r} p(r) \log\left(q(r)\right) \leq \int_{r} p(r) \log\left(p(r)\right) \; , 
\end{align}
where $p(r)$ denotes the (unknown) true distribution of $r$. By inserting \eqref{eq:kl_pos} into~\eqref{eq:kl}, we obtain
\begin{align}
    \text{KL}[p(r \vert x_0, c) \Vert q(r)] &\geq \int_{r,x_0,c} p(r,x_0,c) \log \left(p(r \vert x_0,c) \right) - \int_{r}p(r)\log \left(p(r) \right) \notag \\
                                            & = \int_{r,x_0,c} p(r,x_0,c) \log \left( \frac{p(r \vert x_0,c)}{p(r)} \right) \notag \\
                                            \label{eq:upper_bound}& = \int_{r,x_0,c} p(r,x_0,c) \log \left( \frac{p(r,x_0,c)}{p(r)p(x_0,c)} \right) = \text{MI}\left[r, (x_0,c)\right] \; .
\end{align}
As \eqref{eq:upper_bound} implies, $\text{KL}[p(r \vert x_0, c) \Vert q(r)] \geq \text{MI}\left[r, (x_0,c)\right]$ is an upper bound on the mutual information $\text{MI}\left[r, (x_0,c)\right]$ and, thus, its minimizer also forces independence of $r$ and $(x_0,c)$.\\
\subsection{Derivation of Objective Function}
\label{ap:loss}
\noindent
Again, we start with the definition of the Kullback-Leibler-Divergence and follow~\cite{rombach2_network} 
\begin{align}
    \text{KL}[p(r \vert x_0, c) \Vert q(r)] &= \int_{r} p(r \vert x_0,c) \log \left( \frac{p(r \vert x_0,c)}{q(r)} \right) \notag \\
    \label{eq:dtr}&=\int_{\boldsymbol{X}} p\left(\tau_{\theta}^{-1}(\boldsymbol{X} \vert x_0,c) \vert x_0, c\right) \cdot \vert \det J_{\tau_{\theta}^{-1}}(\boldsymbol{X} \vert x_0, c) \vert  \cdot \log\left( \frac{p\left(\tau_{\theta}^{-1}(\boldsymbol{X} \vert x_0,c)\vert x_0, c\right)}{q\left(\tau_{\theta}^{-1}(\boldsymbol{X} \vert x_0,c)\right)} \right) \; ,
\end{align}
\noindent where we have made a change of integration variable from $r$ to $\boldsymbol{X}$. Further, by rearranging Eq.~\eqref{eq:transformation}, we have 
\begin{equation}
    p(\tau_{\theta}^{-1}(\boldsymbol{X} \vert x_0, c)\vert x_0, c) = \frac{p(\boldsymbol{X} \vert x_0, c)}{\vert \det J_{\tau_{\theta}^{-1}}(\boldsymbol{X} \vert x_0, c) \vert} \;.
    \label{eq:inverse_function}
\end{equation}
Inserting~\eqref{eq:inverse_function} into~\eqref{eq:dtr} yields
\begin{align}
    \text{KL}[p(r \vert x_0, c) \Vert q(r)] &=\small\int_{\boldsymbol{X}} p(\boldsymbol{X} \vert x_0, c) \log \left(\frac{p(\boldsymbol{X} \vert x_0, c)}{q\left(\tau_{\theta}^{-1}(\boldsymbol{X} \vert x_0,c)\right) \vert \det J_{\tau_{\theta}^{-1}}(\boldsymbol{X} \vert x_0, c) \vert}\right)\notag \\
    \label{eq:exp}&= \mathbb{E}_{\boldsymbol{X}}\Big[\log p(\boldsymbol{X} \vert x_0, c) - \log \left( q\left(\tau_{\theta}^{-1}(\boldsymbol{X} \vert x_0,c)\right)\right) - \log \left( \vert \det J_{\tau_{\theta}^{-1}}(\boldsymbol{X} \vert x_0, c) \vert \right)\Big] \;.
\end{align}
If we now choose $q(r) = \mathcal{N}(0,\boldsymbol{I})$ and insert this into~\eqref{eq:exp}, we obtain our final objective by taking the expectation over $(x_0, c)$, as 
\begin{align}
    \text{KL}[p(r \vert x_0, c) \Vert q(r)] &= \mathbb{E}_{\boldsymbol{X},x_0,c}\Big[\frac{1}{2} \Vert \tau_{\theta}^{-1}(\boldsymbol{X} \vert x_0, c) \Vert^{2}_{2}- \log \left( \vert \det J_{\tau_{\theta}^{-1}}(\boldsymbol{X} \vert x_0, c) \vert \right) \Big] + H \notag \\
  & \propto \mathbb{E}_{\boldsymbol{X},x_0, c}\Big[\Vert \tau_{\theta}^{-1}(\boldsymbol{X} \vert x_0, c) \Vert^{2}_{2} - \log \left( \vert \det J_{\tau_{\theta}^{-1}}(\boldsymbol{X} \vert x_0, c) \vert \right) \Big]  \; ,
\end{align}
\noindent where the constant entropy of the data $H = \mathbb{E}_{\boldsymbol{X},x_0, c}\left[\log p(\boldsymbol{X} \vert x_0, c)\right]$ can be neglected for optimizing $\tau_{\theta}$.\\
Summarizing the derived results, we can see, that minimizing $\text{KL}[p(r \vert x_0, c) \Vert q(r)]$ not only yields a probabilistic generative model capturing the data distribution, but also forces independence between the latent variable $r$ and the conditioning $(x_0,c)$. Interestingly, we can compare our model with cVAE-based models~\cite{VAE,alemi2016,rezende2014stochastic,cvae} which have to balance between two contradicting loss terms where one ensures good reconstruction properties and the other induces independence between the latent variable and the conditioning. Thus, these models face the trade-off mentioned in the main paper. As derived above, minimizing $\text{KL}[p(r \vert x_0, c) \Vert q(r)]$ induces both these properties without forcing the model to balance between them (for a broader discussion, see~\cite{rombach2_network}).

\section{Implementation Details of iPOKE}
In this Section, we provide detailed information regarding our employed model architectures and losses as well as specifications of the models' hyperparameters. Our model is implemented using PyTorch~\cite{pytorch}. The corresponding code is available at \url{https://bit.ly/36h8OKr}.
\label{ap:impl_ipoke}
\subsection{Video Autoencoding Framework}
\label{ap:ae}
\noindent
We now provide details regarding the video autoencoding framework which is used to retain tractable computational cost for training our conditionally invertible model $\tau_{\theta}$ (cf Section~3.2 in the main paper). The autoencoding framework is trained prior to optimizing $\tau_{\theta}$. In the following we will first describe how to train this model, before its implementation details are given.\\ 
\noindent
\textbf{Model Overview and Losses.}\\
\noindent \textit{Model Overview.}
As stated in the main paper, we pre-train a video-autoencoder to obtain a compact, information-preserving video encoding $z = E(\boldsymbol{X}) \in \mathbb{R}^{h \times w \times d}$ as fixed input for our invertible model $\tau_{\theta}$, where $E$ is a 3D-ResNet-encoder~\cite{he2015ResNet}. While training the autoencoding framework, $z=\hat{z}_0$ constitutes the initial hidden state of the latent GRU~\cite{gru} predicting the (temporally) subsequent states $\hat{z}_i = \text{GRU}(\hat{z}_{i-1})\; , \; i = 1, \dots T$ which are mapped back into the image domain by using a 2D-decoder, thus yielding the individual images frames $\hat{x}_i = D(z_i)$ of a predicted sequence $\hat{\boldsymbol{X}}$. The autoencoding framework is visualized together with our invertible model in Fig.~\ref{fig:method}.\\
\noindent\textit{Training the Autoencoder.}
Here we explain the individual terms of the overall loss function of this model~\eqref{eq:ae_loss} in more detail. The first term, $\mathcal{L}_{rec}$, is an average reconstruction loss between the individual image frames of the ground-truth and predicted sequences, consisting of an $L1$-loss in the pixel space and a perceptual loss $\ell^{\phi}$~\cite{dosovitskiy201perceptual, johnson2016perceptual}.
\begin{equation}
\label{eq:perc_loss}
\mathcal{L}_{rec} = \frac{1}{T} \left[ \sum_{i=1}^{T}\parallel x_i - \hat{x}_i \parallel_1 +  \sum_{i=1}^{T}\ell^{\phi}(x_i, \hat{x}_i) \right]
\end{equation}
Moreover, we follow previous work~\cite{DVDGAN, vid2vid} and employ a spatial discriminator $\mathcal{D}_S$ and a temporal discriminator $\mathcal{D}_T$. Both discriminators are optimized using the hinge formulation~\cite{lim2017geometric, big_gan_brock}. 
For stabilizing the GAN training, gradient penalty \cite{MeschederICML2018, gulrajani2017improved} is used for the temporal discriminator. Additionally, we add a feature matching loss \cite{high_res_image_syn} for each individual discriminator, thus yielding the adversarial objectives 
\begin{equation}
    \mathcal{L}_{\mathcal{D}_S} =\frac{1}{T} \cdot \sum_{i=1}^{T} \left[ \mathcal{D}_S(\hat{x}_i) - \ell_{F_S}(x_i, \hat{x}_i) \right] \; ,
\label{eq:loss_dstatic}
\end{equation}
and
\begin{equation}
\mathcal{L}_{\mathcal{D}_T} = \mathcal{D}_T(\hat{\boldsymbol{X}}) + \ell_{F_T}(\boldsymbol{X}, \hat{\boldsymbol{X}}) 
\end{equation}
for the generator, where $\ell_{F_T}$ and $\ell_{F_S}$ denote additional feature matching losses~\cite{high_res_image_syn}.\\ 
The counterfighting loss objective for the temporal discriminator can be written as 
\begin{equation}
    \mathcal{L}_{\mathcal{D}_T,\mathcal{D}} =  \rho (1-\mathcal{D}_T(\boldsymbol{X})) + \lambda_{gp} \parallel \nabla \mathcal{D}_T(\boldsymbol{X}) \parallel_2^2  
    + \rho (1 + \mathcal{D}_T(\hat{\boldsymbol{X}})) \;,
\end{equation}
where $\parallel \nabla \mathcal{D}_T(X) \parallel_2^2$ denotes the gradient penalty~\cite{MeschederICML2018, gulrajani2017improved} with $\lambda_{gp}$ as its weighting factor and $\rho$ the ReLU activation function. For the spatial discriminator, the objective can be formulated as 
\begin{equation}
    \mathcal{L}_{\mathcal{D}_S,\mathcal{D}} =\frac{1}{T} \sum_{i=1}^{T} [\rho (1-\mathcal{D}_S(x_i)) + \rho (1 + \mathcal{D}_S(\hat{x_i}))] \;.
\end{equation}\\
\noindent
\textbf{Implementation details.}\\
\noindent \textit{Encoder $E$.} The encoder $E$ is a 3D-ResNet built up of layers following the structure proposed by~\cite{he2015ResNet}. GroupNorm~\cite{wu2018group} is used as a normalization layer. The number of subsequently applied layers varies with the spatial size of the input video sequence $\boldsymbol{X}$, such that a latent encoding $z$ of spatial size $h = w = 8$ is obtained in all cases. The number of channels $d$ of $z$ is chosen as $d = 32$ for the iPER~\cite{iper} and Tai-Chi~\cite{first_order_2019} datasets and $d = 64$ for the PP and Human3.6m~\cite{h36m} datasets.\\
\noindent \textit{GRU.} 
We use a convolutional-GRU with 4 hidden layers. The number of channels remains the same for each hidden layer. Within each GRU cell, we use common convolutions with kernel size $3 \times 3$. As stated above, $z$ initializes the hidden state for all hidden layers of the GRU model. Subsequent hidden states are re-fed into the network until the intended sequence length is obtained. The input state is a learned constant of the same shape than the hidden state for all time steps.\\
\noindent \textit{Decoder $D$.} The decoder is of a sequence of ELU-activated~\cite{elu} 2D-ResNet blocks~\cite{he2015ResNet} consisting of convolutions with $3 \times 3$ kernels, where all but the first ResNet blocks upsample their input feature maps by a factor of two until the intended output resolution is reached. If upsampling is applied, the first convolution and the skip connection of the respective ResNet-block are replaced by a transposed convolution. Spectral normalization is applied to each of the convolutional layers contained in the decoder. Further, we use a SPADE-normalization layer which gets the start frame $x_0$ the respective sequence as input after each upsampling ResNet block.\\
\noindent \textit{Discriminators.}
For the static discriminator, a patch discriminator \cite{pix2pix} is used and for the temporal discriminator a 3D ResNet-18 \cite{he2015ResNet}. Both discriminators use GroupNorm \cite{wu2018group} as a normalization layer.\\
\noindent \textit{Training Details.}
We train our model to reconstruct videos of sequence length $T=10$ comprising images of spatial size $64 \times 64$ for comparison with the stochastic video synthesis baselines and $128 \times 128$ for all other experiments which are considered within the main paper. When training on sequences of spatial size $64 \times 64$, we use batch size 16 on a single NVIDIA RTX 2080 TI, whereas in the latter case, we use a single NIVDIA A100 with batch size 20. In both cases, all hyperparameters which are subsequently reported were also equally. The models are trained using Adam \cite{kingma2017adam} with a learning rate of $2 \cdot 10^{-4}$, $\beta_1=0.5$, $\beta_2 = 0.9$, weight decay of $10^{-5}$, and exponential learning rate decay. The weighting factor of the gradient penalty for $\mathcal{D}_T$ is chosen as $\lambda_{gp} = 1.2$.  

\subsection{Conditional Invertible Neural Network}
\label{ap:cinn}
\noindent
In the following, we provide additional implementation details for our conditional invertible model $\tau_{\theta}$ and the encoders $\Phi_{c}$ and $\Phi_{x_0}$ processing the individual components of $(x_0, c)$.\\
\noindent
\textbf{Encoders $\Phi_{c}$ and $\Phi_{x_{0}}$.}
To obtain descriptive representations capturing all details regarding the two conditioning variables $c$ and $x_0$ we pretrain the respective encoders $\Phi_{c}$ and $\Phi_{x_{0}}$. We train $\Phi_{x_{0}}$ and a subsequent decoder to reconstruct image frames by using a combination of a standard $L1$ reconstruction loss and an additional perceptual loss as a training objective. $\Phi_{c}$ is also combined with a decoder at training time and gets a simulated user control $c$ (cf. Section~\ref{sec:poke_sim}) and the related source image frame $x_0$ as inputs. The obtained model is trained to reconstruct the entire flow field between $x_0$ and $x_T$. Again, a combination of $L1$ and an additional perceptual loss is used as loss function. In both cases, the decoders are used only for training and discarded afterwards. \\
\noindent
\textbf{Conditional INN $\tau_{\theta}$.}\\
\noindent
\textit{Model Architecture.}
The cINN is implemened as a convolutional normalizing flow model~\cite{rezende2016variational,lugmayr2020srflow,glow,papamakarios2019normalizing}. As stated in the main paper, it is built up of $K$ subsequently arranged flow blocks, where the $k$-th block comprises $N_k$ sub-blocks, followed by a modified glow block~\cite{glow}, consisting of ActNorm~\cite{glow}, affine coupling~\cite{dinh2017density} and shuffling layers\footnote{We modify these blocks by using shuffling layers instead of invertible $1\times1$ convolutions, as originally proposed in glow~\cite{glow}.}. In each sub=block, we employ masked convolutions~\cite{macow_2019}, which subdivide a convolutional kernel into four systematically rotated, efficiently invertible sub-kernels, whose Jacobian is tractable. These properties are crucial when training normalizing flow model~\cite{rezende2016variational,lugmayr2020srflow,glow,papamakarios2019normalizing}, as they guarantee tractable training times and efficient sampling from $q(r)$. Moreover, masked convolutions result in the same receptive field than common convolutions and, thus, show improved expressivity compared to widely used architectures such as affine coupling layers~\cite{dinh2015nice,dinh2017density}. The flow overall architecture and the sub-blocks are visualized in  Figure~\ref{fig:method} in the main paper. For more details regarding masked convolutions, we refer the reader to~\cite{macow_2019}.\\
Within all models and datasets , we use $K =15$ flow blocks. The number $N_k$ of sub-blocks contained in the the $k$-th flow block decreases with increasing $k$. The individual number of sub blocks per flow block can be summarized by the following vector
\begin{equation*}
    \boldsymbol{N}_{1:K} = [10,5,5,4,4,4,3,3,3,2,2,2,1,1,1]
\end{equation*}
where the $k$-th component of $\boldsymbol{N}_{1:K}$ denotes the number of sub-blocks within flow block $k$.\\ 
\noindent 
\textit{Training Details.}
We train our cINN model using Adam~\cite{adam} optimizer with parameters $\beta_1=0.9$, $\beta_2 = 0.999$ and weight decay of $10^{-5}$. The learning rate is annealed from zero to $10^{-3}$ for the first $500$ train steps and linearly decrease afterwards, until training ends. We train the cINN models on a single NVIDIA A100 using batches comprising $40$ individual training examples.

\subsection{Further Training Details of iPOKE}
\label{ap:further_details}
\noindent
\textbf{Foreground-Background-Separation}
We do not only train our model to infer the implications of a localized poke onto the entire object, we also intend to learn iPOKE such that it only generates object kinematics for control inputs actually acting at the object itself. Hence a poking of a pixel in background areas unrelated to the object should be ignored. We now present an additional training strategy ensuring this property.\\
We assume parts of the background of the videos within the train set to be static and the foreground to obtain a sufficient amount of motion, indicated by a specific magnitude of optical flow. Thus, during training, we only consider those locations with an optical flow magnitude larger than the mean of magnitudes of the flow map $F$ for sampling the controlled location $l = c_{3:4}$ for a user control with the depicted object. However, as we also want our model to separate the pixels on the object surface from those in the background, we sample the twelfth of all controlled locations in each epoch out of background pixels and construct artificial user controls by sampling the magnitudes and angles at these locations from the locations within the foreground. If our model gets such an artificial poke as input, it is trained to reconstruct a still sequence obtained by repeating the source image $x_0$ $T$ times. This strategy is applied both to train the autoencoding framework described in Sec.~\ref{ap:ae} and the cINN model $\tau_{\theta}$. Thus, the model learns to separate the pixels comprising the object from those in the background, as indicated by the video examples in Section~\ref{ap:human}, where we show examples for such artificial user controls.\\

\section{Implementation Details of Baselines}
\label{ap:impl_baselines}
\noindent
\textbf{Video Prediction Models.} 
For comparison, we implemented the competing stochastic video prediction baselines~\cite{2018savp,vrnn-hier,Franceschi2020} based on the officially provided code from Github~\footnote{https://github.com/edouardelasalles/srvp}\footnote{https://github.com/facebookresearch/improved\_vrnn}\footnote{\label{savp_code}https://github.com/alexlee-gk/video\_prediction}. As no pretrained models are available for the utilized datasets, we trained models from scratch for all competitors except for SRVP~\cite{Franceschi2020}, which provide a pretrained model for the Human3.6m~\cite{h36m} dataset. All models are trained to predict sequences of length 10 and spatial size $64 \times 64$ based on two context frames. For models trained from scratch, we used the hyperparameters proposed in the respective publications. For SAVP~\cite{2018savp} we further tried all individual hyperparameter-settings which are available in the official code repository\textsuperscript{\ref{savp_code}}, but could not manage to obtain any visual diversity.\\
\noindent 
\textbf{Controlled Video Synthesis Models.}
Both controllable video synthesis baselines we compare iPOKE with~\cite{controllable_image, poke_blattmann_2021} are implemented based on the official code repositories\footnote{https://github.com/zekunhao1995/ControllableVideoGen}\footnote{https://github.com/CompVis/interactive-image2video-synthesis} and the provided hyperparameters for generating videos of spatial size $128 \times 128$ on all used datasets. For \cite{controllable_image}, due to the lack of pretrained weights, we trained the new models on the considered datasets, contrasting II2V~\cite{poke_blattmann_2021}, where we entirely used the available pretrained models.\\
For the method of Hao et al~\cite{controllable_image} we follow their proposed procedure to predict videos $[x_0,\dots,x_T]$ of length $T$: we first sample $k$ flow vectors between $x_0$ and $x_1$. We then extend these initial vectors to discrete trajectories of length $T-1$ in the image space by tracking the initial points using consecutive flow vectors between $x_{i}$ and $x_{i+1}$. Videos are finally obtained by successively warping $x_0$ using the shift vectors from the trajectories. We follow the official protocol from ~\cite{controllable_image} and set $k=5$. The motion trajectories are generated from the same optical flow which was used to train our own model.\\
For the model trained on ground truth keypoints on the iPER dataset, we constructed the motion trajectories based on the shifts between the individual keypoints of the source frame and all remaining frames of the considered video sequence.\\ 
\noindent
\textbf{Variational Counterpart of iPOKE.}
We here report the implementation details of the cVAE-based counterpart of iPOKE, which we compared our model to in the ablation study in Section~\ref{sec:eval_quan}. The model is composed of the individual parts of the autoencoding framework of iPOKE as described in Sec.~\ref{ap:ae}. Thus, the cVAE-grounded baseline consists of a 3D-Encoder, a convolutional GRU and a 2D-decoder following the respective architectures presented in Section~\ref{ap:ae}. Additionally, we use two encoders to individually encode the components of the conditioning $(x_0, c)$ with the same architecture than $\Phi_{x_{0}}$ and $\Phi_c$ as described in Section~\ref{ap:cinn}. All these components of the overall model have the same inputs and outputs as the corresponding components of iPOKE. The differences between the two models are \textit{i)} the regularization of the video encoding $z$ towards the standard normal prior by using the reparameterization trick~\cite{VAE}, as usually done in cVAEs~\cite{VAE,rezende2014stochastic}, instead of using a cINN operating on the latent level, and \textit{ii)} the initialization of the hidden state of the latent GRU, which are here the stacked features of the reparametrized video encoding and the representations of the source image $x_0$ and the user control $c$. For iPOKE, in contrast, the initial hidden state is only the video representation $z = \tau_{\theta}(r \vert x_0, c)$.\\
The losses for the model are exactly similar to those of our video autoencoding framework as summarized in Eq.~\eqref{eq:ae_loss}, except for an additional KL-loss ensuring the mentioned regularization of the video representation towards the standard normal. We tried different options for the weighting factor $\lambda_{KL}$ of this KL-term, as its choice has a major influence on the performance of the trained model and, thus, needs to be selected carefully~\cite{chen2016variational, zhao2017infovae}. The best performance was obtained by choosing $\lambda_{KL} = 0.1$. Additionally, to avoid posterior collapse~\cite{Lucas2019UnderstandingPC}, we use KL-annealing for 5 epochs, before the final value of $\lambda_{KL}$ is reached. Similar to our proposed autoencoding model for iPOKE, the VAE-baseline is trained using Adam~\cite{adam} with a learning rate of $2 \cdot 10^{-4}$, $\beta_1=0.5$, $\beta_2 = 0.9$, weight decay of $10^{-5}$, and exponential learning rate decay. All remaining hyperparameters were chosen to be equal to those of the corresponding iPOKE-model for video sequences of spatial size $64 \times 64$. 


\section{Evaluation Details}
\label{ap:eval}
\noindent \textbf{Motion Consistency.}
To compute the FVD-score~\cite{FVD} for a given model, we generated 1000 video sequences and sampled 1000 random videos of the same length from the ground truth data. Both the real and the generated examples are the input to the I3D~\cite{i3d} model which was pretrained on the Kinetics~\cite{kinetics} dataset. Subsequently their distributions in the I3D feature space are compared resulting in the reported FVD-scores. For all models we concatenate the last conditioning frame (for our model the input frame) with the generated sequence. \\
\noindent \textbf{Synthesis Quality.}
All reported accuracy metrics are based on 1000 predicted video sequences and the corresponding ground truth videos. As these metrics are calculated based on individual image frames, we compare each frame of a generated sequence with its corresponding frame in the ground truth sequence, resulting in $T$ scores for a predicted video of length $T$, which are subsequently averaged to obtain a scalar value per sequence.  
\\
\noindent \textbf{Motion Diversity.}
To evaluate the diversity for each model, we generated 5 realizations for each video resulting in $5\times 1000$ videos. The diversity is then computed between the different realizations of the video using LPIPS\cite{lpips} and MSE in the pixel space. We did not notice any change in numbers when using more realizations.

\clearpage
{\small
\bibliographystyle{ieee_fullname}
\bibliography{egbib}
}

\end{document}